\pdfoutput=1

\documentclass[11pt]{article}

\usepackage{acl2023}

\usepackage{times}
\usepackage{latexsym}

\usepackage[T1]{fontenc}

\usepackage[utf8]{inputenc}

\usepackage{microtype}

\usepackage{inconsolata}

\usepackage{times}
\usepackage{latexsym}
\usepackage{color}
\usepackage{xcolor}
\usepackage{booktabs}
\usepackage{multirow}
\usepackage{xspace}
\usepackage{graphicx}
\usepackage{todonotes}
\usepackage{enumitem}

\title{GENEVA: Benchmarking Generalizability for Event Argument Extraction with Hundreds of Event Types and Argument Roles}

\author{Tanmay Parekh$^{\dagger}$ \ \ \
I-Hung Hsu$^{\ddagger}$ \ \ \
Kuan-Hao Huang$^{\dagger}$ \\
{\bf Kai-Wei Chang$^{\dagger}$ \ \ \
Nanyun Peng$^{\dagger}$} \\
$^{\dagger}$Computer Science Department, University of California, Los Angeles \\
$^{\ddagger}$Information Science Institute, University of Southern California \\
\texttt{\{tparekh, khhuang, kwchang, violetpeng\}@cs.ucla.edu} \\
\texttt{\{ihunghsu\}@isi.edu}
  }

\newcommand{\dataName}[0]{GENEVA}
\newcommand{\modelName}[0]{DEGREE}
\newcommand{\degree}[0]{\textsc{Degree}}

\newcommand{\mypar}[1]{\vspace{0.35em}\noindent\textbf{#1}}

\begin{document}
\maketitle
\begin{abstract}

Recent works in Event Argument Extraction (EAE) have focused on improving model generalizability to cater to new events and domains.
However, standard benchmarking datasets like ACE and ERE cover less than $40$ event types and $25$ entity-centric argument roles.
Limited diversity and coverage hinder these datasets from adequately evaluating the generalizability of EAE models.
In this paper, we first contribute by creating a large and diverse EAE ontology.
This ontology is created by transforming FrameNet, a comprehensive semantic role labeling (SRL) dataset for EAE, by exploiting the similarity between these two tasks. 
Then, exhaustive human expert annotations are collected to build the ontology, concluding with $115$ events and $220$ argument roles, with a significant portion of roles not being entities.
We utilize this ontology to further introduce \dataName{}, a diverse generalizability benchmarking dataset comprising four test suites, aimed at evaluating models' ability to handle limited data and unseen event type generalization.
We benchmark six EAE models from various families.
The results show that owing to non-entity argument roles, even the best-performing model can only achieve $39\%$ F1 score,
indicating how \dataName{} provides new challenges for generalization in EAE.
Overall, our large and diverse EAE ontology can aid in creating more comprehensive future resources, while \dataName{} is a challenging benchmarking dataset encouraging further research for improving generalizability in EAE. The code and data can be found at \url{https://github.com/PlusLabNLP/GENEVA}.
\looseness=-1

\end{abstract}

\section{Introduction}

\begin{figure}[t]
    \centering
    \includegraphics[width=0.48\textwidth]{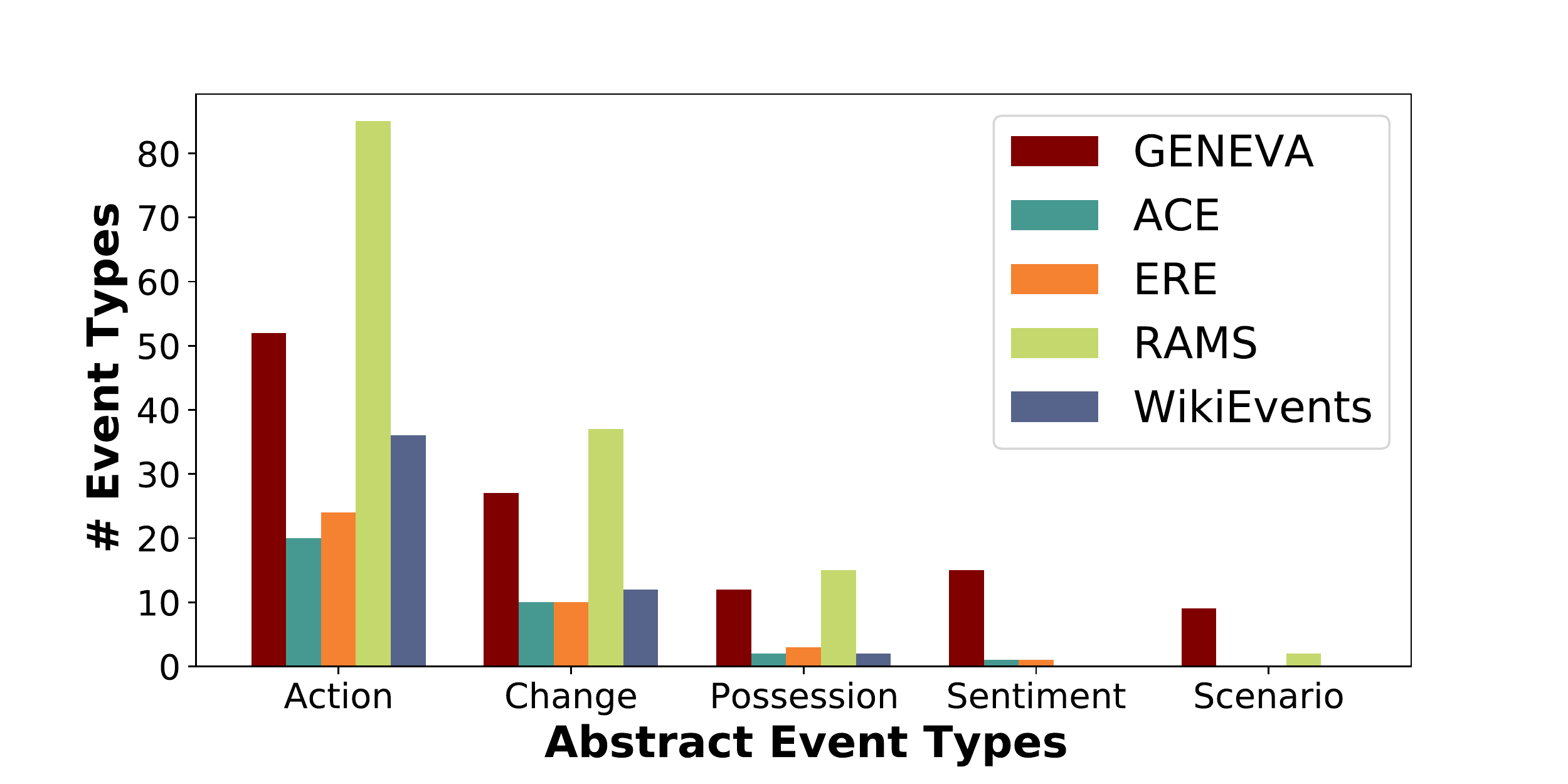}
    \caption{Distribution of event types into various abstract event types$^1$ for \dataName{}, ACE, ERE, RAMS, and WikiEvents datasets.
    We observe that \dataName{} is relatively more diverse than the other datasets.}
    \label{fig:ace-in-geneva}
    \vspace{-1em}
\end{figure}

Event Argument Extraction (EAE) aims at extracting structured information of event-specific arguments and their roles for events from a pre-defined taxonomy.
EAE is a classic topic \cite{sundheim-1992-overview} and elemental for a wide range of applications like building knowledge graphs \cite{10.1145/3366423.3380107}, question answering \cite{berant-etal-2014-modeling}, and others \cite{hogenboom2016survey, DBLP:conf/naacl/WenLLPLLZLWZYDW21, yang-etal-2019-interpretable}.
Recent works have focused on building generalizable EAE models \cite{huang-etal-2018-zero, lyu-etal-2021-zero, sainz-etal-2022-textual} and they utilize existing datasets like ACE \cite{doddington-etal-2004-automatic} and ERE \cite{DBLP:conf/aclevents/SongBSRMEWKRM15} for benchmarking.
However, as shown in Figure~\ref{fig:ace-in-geneva}, these datasets have limited diversity as they focus only on two abstract types,\footnote{Abstract event types are defined as the top nodes of the event ontology created by MAVEN \cite{wang-etal-2020-maven}.} Action and Change.
Furthermore, they have restricted coverage as they only comprise argument roles that are entities.
The limited diversity and coverage restrict the ability of these existing datasets to robustly evaluate the generalizability of EAE models.
Toward this end, we propose a new generalizability benchmarking dataset in our work.

To build a strong comprehensive benchmarking dataset, we first create a large and diverse ontology.
Creating such an ontology from scratch is time-consuming and requires expert knowledge.
To reduce human effort, we exploit the shared properties between semantic role labeling (SRL) and EAE \cite{aguilar-etal-2014-comparison} and leverage a diverse and exhaustive SRL dataset, FrameNet \cite{baker-etal-1998-berkeley-framenet}, to build the ontology.
Through extensive human expert annotations, we design mappings that transform the FrameNet schema to a large and diverse EAE ontology, spanning $115$ event types from five different abstract types.
Our ontology is also comprehensive, comprising $220$ argument roles with a significant $37\%$ of roles as non-entities.


Utilizing this ontology, we create \dataName{} - a \textbf{G}eneralizability B\textbf{EN}chmarking Dataset for \textbf{EV}ent \textbf{A}rgument Extraction.
We exploit the human-curated ontology mappings to transfer FrameNet data for EAE to build \dataName{}.
We further perform several human validation assessments to ensure high annotation quality.
\dataName{} comprises four test suites to assess the models' ability to learn from limited training data and generalize to unseen event types.
These test suites are distinctly different based on the training and test data creation --
(1) low resource, (2) few-shot, (3) zero-shot, and (4) cross-type transfer settings.

We use these test suites to benchmark various classes of EAE models - 
traditional classification-based models \cite{wadden-etal-2019-entity, lin-etal-2020-joint, wang-etal-2022-query}, question-answering-based models \cite{du-cardie-2020-event}, and generative approaches \cite{Paolini21tacl, hsu2021degree}.
We also introduce new automated refinements in the low resource state-of-the-art model DEGREE \cite{hsu2021degree} to generalize and scale up its manual input prompts.
Experiments reveal that DEGREE performs the best and exhibits the best generalizability.
However, owing to non-entity arguments in \dataName{}, DEGREE achieves an F1 score of only $39\%$ on the zero-shot suite.
Under a similar setup on ACE, DEGREE achieves $53\%$, indicating how \dataName{} poses additional challenges for generalizability benchmarking.


To summarize, we make the following contributions.
We construct a diverse and comprehensive EAE ontology introducing non-entity argument roles. This ontology can be utilized further to develop more comprehensive datasets for EAE.
In addition, we propose a generalizability evaluation dataset \dataName{} and benchmark various recent EAE models.
Finally, we show how \dataName{} is a challenging dataset, thus, encouraging future research for generalization in EAE.

\section{Related Work}

\mypar{Event Extraction Datasets and Ontologies:}
The earliest datasets in event extraction date back to MUC \cite{sundheim-1992-overview, grishman-sundheim-1996-message}.
\citet{doddington-etal-2004-automatic} introduced the standard dataset ACE while restricting the ontology to focus on entity-centric arguments.
The ACE ontology was further simplified and extended to ERE \cite{DBLP:conf/aclevents/SongBSRMEWKRM15}
and various TAC KBP Challenges \cite{ellis2014overview, ellis2015overview, getman2017overview}.
These datasets cover a small and restricted set of event types and argument roles with limited diversity.
Later, MAVEN \cite{wang-etal-2020-maven} introduced a massive dataset spanning a wide range of event types.
However, its ontology is limited to the task of Event Detection\footnote{Event Detection aims at only identifying the event type documented in the sentence.} and does not contain argument roles.
Recent works have introduced document-level EAE datasets like RAMS \cite{ebner-etal-2020-multi}, WikiEvents \cite{li-etal-2021-document}, and DocEE \cite{tong-etal-2022-docee}; but their ontologies are also entity-centric, and their event coverage is limited to specific abstract event types (Figure~\ref{fig:ace-in-geneva}).
In our work, we focus on building a diverse and comprehensive dataset for benchmarking generalizability for sentence-level EAE.

\mypar{Event Argument Extraction Models:}
Traditionally, EAE has been formulated as a classification problem \cite{nguyen-etal-2016-joint-event}.
Previous classification-based approaches have utilized pipelined approaches \cite{yang-etal-2019-exploring, wadden-etal-2019-entity} as well as incorporating global features for joint inference \cite{li-etal-2013-joint, yang-mitchell-2016-joint, lin-etal-2020-joint}.
However, these approaches exhibit poor generalizability in the low-data setting \cite{liu-etal-2020-event, hsu2021degree}.
To improve generalizability, some works have explored better usage of label semantics by formulating EAE as a question-answering task \cite{liu-etal-2020-event, li-etal-2020-event, du-cardie-2020-event}.
Recent approaches have explored the use of natural language generative models for structured prediction to boost generalizability \cite{schick-schutze-2021-exploiting, schick-schutze-2021-just, Paolini21tacl, li-etal-2021-document}.
Another set of works transfers knowledge from similar tasks like abstract meaning representation and semantic role labeling \cite{huang-etal-2018-zero, lyu-etal-2021-zero, zhang-etal-2021-zero}.
\degree~\cite{hsu2021degree} is a recently introduced state-of-the-art generative model which has shown the best performance in the limited data regime.
In our work, we benchmark the generalizability of various classes of old and new models on our dataset. \looseness=-1



\section{Ontology Creation}

\vspace{-1mm}

Event annotations start with ontology creation, which defines the scope of the events and their corresponding argument roles of interests.
Towards this end, we aim to construct a large ontology of diverse event types with an exhaustive set of event argument roles.
However, it is a challenging and tedious task that requires extensive expert supervision if building from scratch. 
To reduce human effort while maintaining high quality, we leverage the shared properties of SRL and EAE and utilize a diverse and comprehensive SRL dataset --- FrameNet to design our ontology.
We first re-iterate the EAE terminologies we follow  (\S~\ref{sec:task-definition}) and then describe how FrameNet aids our ontology design (\S~\ref{sec:framenet-for-eae}). 
Finally, we present our steps for creating the final ontology in \S~\ref{sec:event-arg-building} and ontology statistics in \S~\ref{sec:ontology-statistics}.

\vspace{-1mm}

\subsection{Task Definition}
\label{sec:task-definition}

\begin{figure}[t]
    \centering
    \includegraphics[width=0.4\textwidth]{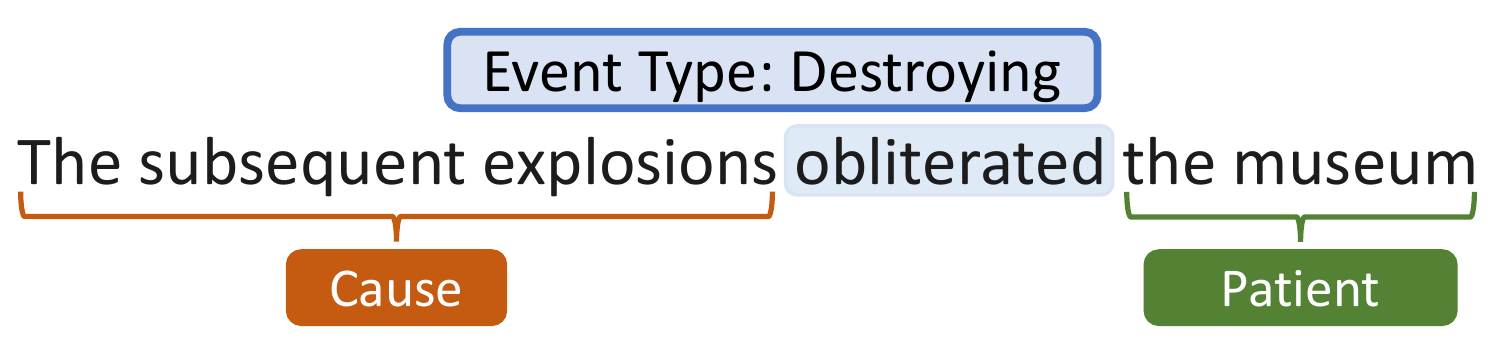}
    \caption{An illustration of EAE for the Destroying event comprising argument roles of Cause and Patient.
    }
    \label{fig:ee_example}
    \vspace{-3mm}
\end{figure}

We follow the definition of \textbf{event} as a class attribute with values such as \textit{occurrence, state, or reporting}~\cite{pustejovsky2003timebank, DBLP:conf/emnlp/HanHSBNRP21}.
\textbf{Event Triggers} are word phrases that best express the occurrence of an event in a sentence.
Following the early works of MUC \cite{sundheim-1992-overview, grishman-sundheim-1996-message}, \textbf{event arguments} are defined as participants in the event which provide specific and salient information about the event. \textbf{Event argument role} is the semantic category of the information the event argument provides.
We provide an illustration in Figure~\ref{fig:ee_example} describing an event about \textit{``Destroying''}, where the event trigger is \textit{obliterated}, and the event consists of argument roles --- \textit{Cause} and \textit{Patient}.


It is worth mentioning that these definitions are disparate from the ones that previous works like ACE, and its inheritors, ERE and RAMS, follow.
In ACE, the scope of events is restricted to the attribute of occurrence only, and event arguments are restricted to entities, wherein \textbf{entities} are defined as objects in the world.
For example, in Figure~\ref{fig:ee_example}, \textit{the subsequent explosions} isn't an entity and will not be considered an argument as per ACE definitions. Consequently, \textit{Cause} won't be part of their ontology.
This exclusion of non-entities leads to incomplete information extraction of the event.
In our work, we follow MUC to consider a broader range of events and event arguments.


\subsection{FrameNet for EAE}
\label{sec:framenet-for-eae}

To overcome the challenge of constructing an event ontology from scratch, we aim to leverage FrameNet, a semantic role labeling (SRL) dataset, to help our ontology creation.
The similarity between SRL and EAE \cite{aguilar-etal-2014-comparison} provides us with the ground for leveraging FrameNet. SRL assigns semantic roles to phrases in the sentence, while EAE extracts event-specific arguments and their roles from the sentence.
Hence, \textit{selecting event-related parts} of a fine-grained annotated SRL dataset can be considered as an exhaustively annotated resource for EAE.

We choose FrameNet\footnote{FrameNet Data Release 1.7 by \url{http://framenet.icsi.berkeley.edu} is licensed under a Creative Commons Attribution 3.0 Unported License.}~\cite{baker-etal-1998-berkeley-framenet} as the auxiliary SRL dataset since it is one of the most comprehensive SRL resources. It comprises 1200+ semantic frames \cite{fillmore1976frame}, where a \textbf{frame} is a holistic background that unites similar words.
Each frame is composed of frame-specific semantic roles (\textbf{frame elements}) and is evoked by specific sets of words (\textbf{lexical units}).

\begin{figure}[t]
    \centering
    \includegraphics[width=0.48\textwidth]{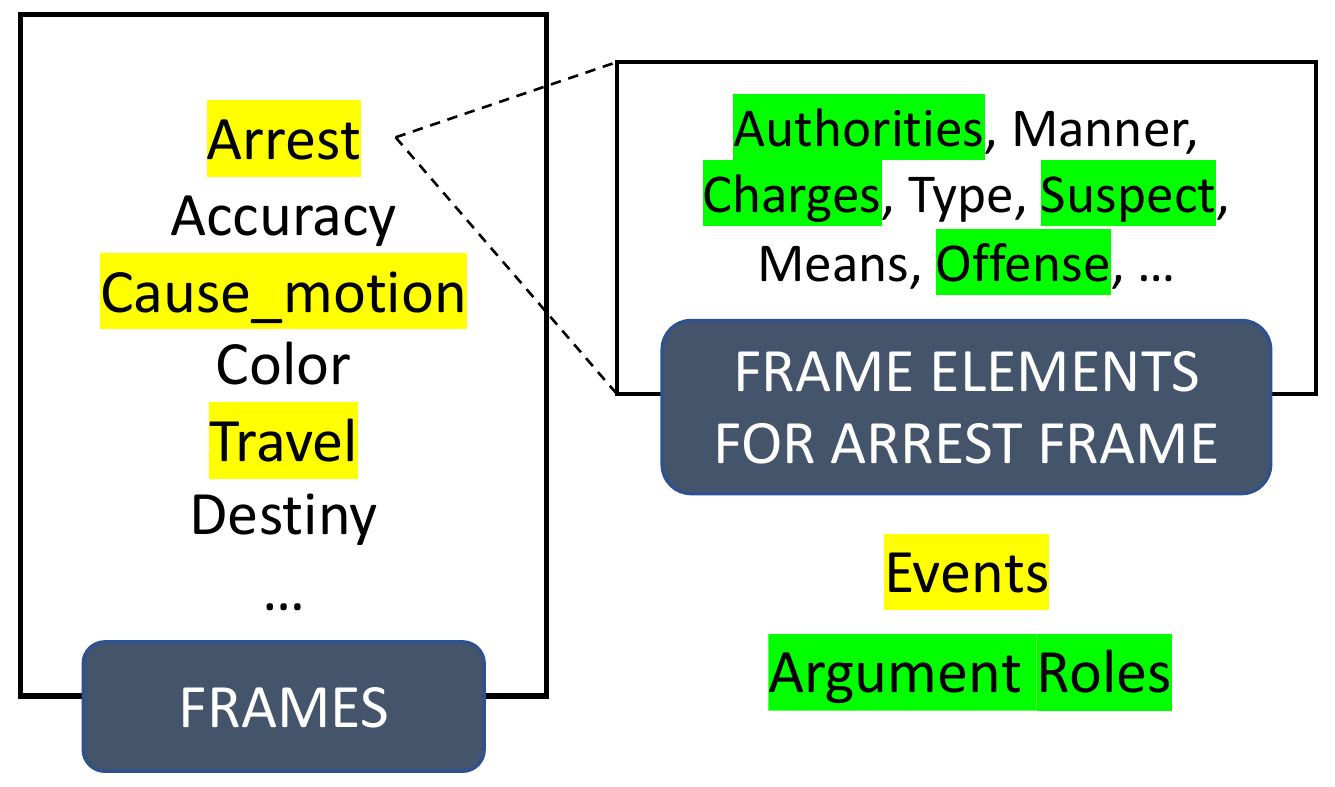}
    \caption{Illustration of challenges in using FrameNet for EAE - Not all frames are events and not all frame elements are argument roles.}
    \label{fig:framenet-not-eae}
    \vspace{-3mm}
\end{figure}

To transfer FrameNet's schema into an EAE ontology, we map \textit{frames} as events, \textit{lexical units} as event triggers, and \textit{frame elements} as argument roles.
However, this basic mapping is inaccurate and has shortcomings since \textit{not all frames are events}, and \textit{not all frame elements are argument roles} per the definitions in \S~\ref{sec:task-definition}.
We highlight these shortcomings in Figure~\ref{fig:framenet-not-eae}, which enlists some FrameNet frames and frame elements for the \textit{Arrest} frame. Based on EAE definitions, only some frames like \textit{Arrest, Travel, etc} (highlighted in yellow) can be mapped as events, and similarly, limited frame elements like \textit{Authorities, Charges, etc} (highlighted in green) are mappable as argument roles.
\looseness=-1

\subsection{Building the EAE Ontology}
\label{sec:event-arg-building}
\begin{figure}[t]
    \centering
    \includegraphics[width=0.48\textwidth]{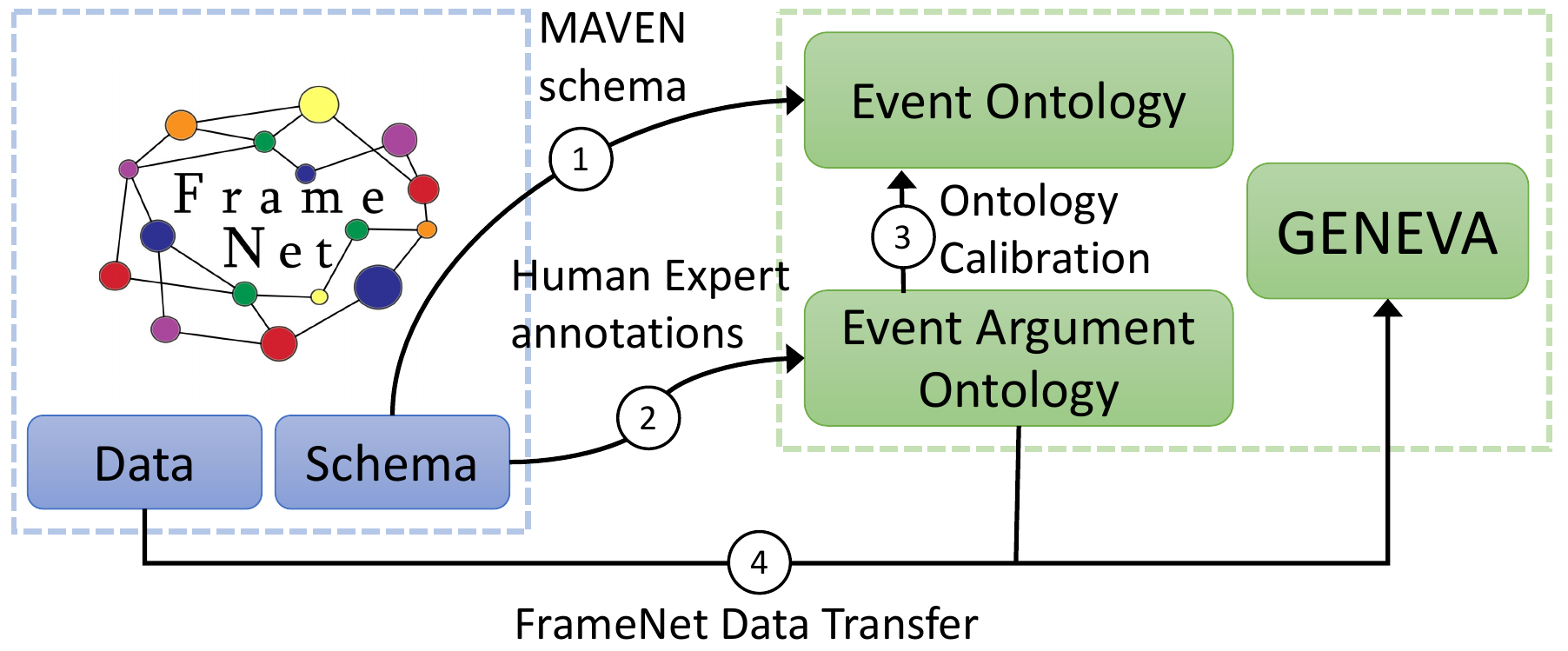}
    \caption{Illustration of the \dataName{} creation from FrameNet labeled sequentially by the crucial steps.
    }
    \label{fig:dataset-creation}
    \vspace{-3mm}
\end{figure}

To overcome the shortcomings of the basic mapping, we follow a two-step approach (Figure~\ref{fig:dataset-creation}).
First, we build an event ontology for accurately mapping frames to events.
Then, we augment this ontology with argument roles by building an event argument ontology.
We describe these steps below.


\mypar{Event Ontology:}
In order to build the event ontology, we utilize the event mapping designed by MAVEN \cite{wang-etal-2020-maven}, which is an event detection dataset.
They first recursively filter frames having a relation with the "Event" frame in FrameNet.
Then they manually filter and merge frames based on the definitions, resulting in an event ontology comprising 168 event types mapped from 289 filtered frames.

\mypar{Event Argument Ontology:}
In order to augment argument roles to the event ontology, we perform an extensive human expert annotation process.
The goal of this annotation process is to create an argument mapping from FrameNet to our ontology by filtering and merging frame elements. We describe this annotation process below.

\noindent \textit{Annotation Instructions:}
Annotators are provided with a list of frame elements along with their descriptions for each frame in the event ontology.~\footnote{Event ontology frames can be viewed as candidate events.}
They are also provided with definitions for events and argument roles as discussed in Section~\ref{sec:task-definition}.
Based on these definitions, they are asked to annotate each frame element as (a) not argument role, (b) argument role, or (c) merge with existing argument role (and mention the argument role to merge with).
To ensure arguments are salient, annotators are instructed to filter out frame elements that are super generic (e.g. Time, Place, Purpose) unless they are relevant to the event.
Ambiguous cases are flagged and commonly reviewed at a later stage.

Additionally, annotators are asked to classify each argument role as an entity or not.
This additional annotation provides flexibility for quick conversion of the ontology to ACE definitions.
Figure~\ref{fig:ontology-annotation} in the Appendix provides an illustration of these instructions and the annotation process.

\noindent \textit{Annotation Results:}
We recruit two human experts who are well-versed in the field of event extraction.
We conduct three rounds of annotations and discussions to improve consistency and ensure a high inter-annotator agreement (IAA).
The final IAA measured as Cohen's Kappa \cite{mchugh2012interrater} was $0.82$ for mapping frame elements and $0.94$ for entity classification.
A total of $3,729$ frame elements from $289$ frames were examined as part of the annotation process.
About $63\%$ frame elements were filtered out, $14\%$ were merged, and the remaining $23\%$ constitute as argument roles.

\mypar{Event Ontology Calibration:}
The MAVEN event ontology is created independent of the argument roles.
This leads to some inaccuracies in their ontology wherein two frames with disparate sets of argument roles are mapped as a single event.
For example, \textit{Surrendering\_possession} and \textit{Surrendering} frames are merged together despite having different argument roles.
Based on our human expert-curated event argument ontology, we rectify these inaccuracies (roughly $8\%$ of the event ontology) and create our final ontology.


\subsection{Ontology Statistics}
\label{sec:ontology-statistics}

We present the statistics of our full ontology in Table~\ref{tab:ontology-statistics} and compare it with existing ACE \cite{doddington-etal-2004-automatic} and RAMS \cite{ebner-etal-2020-multi} ontologies.
But as we will specify in \S~\ref{sec:data-creation}, we use a subset of this ontology\footnote{We will release both full and \dataName{} ontologies to facilitate future study.} for creating \dataName{}.
Hence, we also include the statistics of the \dataName{} ontology in the last column in Table~\ref{tab:ontology-statistics}.
Overall, our curated full ontology is the largest and most comprehensive as it comprises 179 event types and 362 argument roles.
Defining \textit{abstract event types} as the top nodes of the ontology tree created by MAVEN \cite{wang-etal-2020-maven}, we show that our ontology spans $5$ different abstract types and is the most diverse.
We organize our ontology into a hierarchy of these abstract event types in Appendix~\ref{sec:event-organization}.
Our ontology is also dense with an average of $4.82$ argument roles per event type.
Finally, we note that a significant $35\%$ of the event argument roles in our ontology are non-entities.
This demonstrates how our ontology covers a broader and more comprehensive range of argument roles than other ontologies following ACE definitions of entity-centric argument roles.


\begin{table}[t]
    \centering
    \small
    \setlength{\tabcolsep}{3pt}
    \begin{tabular}{lrrr|r}
        \toprule 
         & \textbf{ACE} & \textbf{RAMS} & \textbf{Full} & \textbf{GENEVA} \\
        \midrule
        \# Event Types & $33$ & $139$ & $179$ & $115$ \\
        \# Abstract Event Types & $2$ & $3$ & $5$ & $5$ \\
        \# Argument Roles (AR) & $22$ & $65$ & $362$ & $220$ \\
        Avg. \# AR per Event & $4.75$ & $3.76$ & $4.82$ & $3.97$ \\
        \% Entity AR & $100\%$ & $100\%$ & $65\%$ & $63\%$ \\
        \% Non-Entity AR & $0\%$ & $0\%$ & $35\%$ & $37\%$ \\
        \bottomrule
    \end{tabular}
    \caption{Full and \dataName{} ontology Statistics. AR = Argument Role. An ontology covers an abstract type if it has $5+$ events of that abstract type. Entity AR refers to argument roles that are entities.}
    \label{tab:ontology-statistics}
    \vspace{-3mm}
\end{table}



\section{\dataName{} Dataset}
\label{sec:geneva-dataset}

Previous EAE datasets for evaluating generalizability like ACE and ERE have limited event diversity and are restricted to entity-centric arguments.
To overcome these issues, we utilize our ontology to construct a new generalizability benchmarking dataset \dataName{} comprising four specialized test suites.
We describe our data creation process in \S~\ref{sec:data-creation}, provide data statistics in \S~\ref{sec:data-analysis} and discuss out test suites in \S~\ref{sec:benchmark-setup}.


\subsection{Creation of \dataName}
\label{sec:data-creation}

Since annotating EAE data for our large ontology is an expensive process, we leverage the annotated dataset of FrameNet to create \dataName{} (Figure~\ref{fig:dataset-creation}).
We utilize the previously designed ontology mappings to repurpose the annotated sentences from FrameNet for EAE by mapping frames to corresponding events, lexical units to event triggers, and frame elements to corresponding arguments.
Unmapped frames and frame elements (not in the ontology) are filtered out from the dataset.
Since FrameNet doesn't provide annotations for all frames, some events from the full ontology are not present in our dataset (e.g. \textit{Military\_Operation}).
Additionally, to aid better evaluation, we
remove events that have less than $5$ event mentions (e.g. \textit{Lighting}).
Finally, \dataName{} comprises  115 event types and 220 argument roles.
Some examples are provided in Figure~\ref{fig:dataset-examples} (Appendix).

\mypar{Human Validation:}
We ensure the high quality of our dataset by conducting two human assessments:

\noindent (1) \textit{Ontology Quality Assessment}:
We present the human annotators with three sentences - one primary and two candidates - and ask them if the event in the primary sentence is similar to the events in either of the candidates or distinct from both (Example in Appendix~\ref{sec:human-validation-setup}).
One candidate sentence is chosen from the frame merged with the primary event, while the other candidate is chosen from a similar unmerged sister frame.
The annotators chose the merged frame candidates $87\%$ of the times, demonstrating the high quality of the ontology mappings.
This validation was done by three annotators over $61$ triplets with $0.7$ IAA measured by Fleiss' kappa \cite{fleiss1971measuring}.

\noindent (2) \textit{Annotation Comprehensiveness Assessment}:
Human annotators are presented with annotated samples from our dataset and they are asked to report if there are any arguments in the sentence that have not been annotated.
The annotation is considered comprehensive if all arguments are annotated correctly.
The annotators reported that the annotations were $89\%$ comprehensive, ensuring high dataset quality.
Corrections majorly comprise ambiguous cases and incorrect role labels.
This assessment was done by two experts over $100$ sampled annotations with $0.93$ IAA (Cohen's kappa).

\begin{table}[t]
    \centering
    \small
    \begin{tabular}{lrrrr}
    \toprule
        \multirow{2}{*}{\textbf{Dataset}} & \textbf{\#Event} & \textbf{\#Arg} & \textbf{Avg. Event} & \textbf{Avg. Arg} \\
        & \textbf{Types} & \textbf{Types} & \textbf{Mentions} & \textbf{Mentions} \\
    \midrule
        ACE & 33 & 22 & 153.18 & 274.55 \\
        ERE & 38 & 21 & 191.76 & 499 \\
        \dataName & 115 & 220 & 65.26 & 55.77 \\
    \bottomrule
    \end{tabular}
    \caption{
    \looseness=-1
    Statistics for different EAE datasets for benchmarking generalizability. The second and third columns are the unique number of event types and argument roles. The last two columns indicate the average number of mentions per event and argument role.}
    \label{tab:dataset-coverage}
    \vspace{-3mm}
\end{table}

\subsection{Data Analysis}
\label{sec:data-analysis}

Overall, \dataName{} is a dense, challenging, and diverse EAE dataset with good coverage.
These characteristics make \dataName{} better-suited than existing datasets like ACE/ERE for evaluating the generalizability of EAE models. 
The major statistics for \dataName{} are shown in Table~\ref{tab:dataset-coverage} along with its comparison with ACE and ERE.
We provide more discussions about the characteristics of our dataset as follows.

\noindent \textbf{Diverse:}
\dataName{} has wide coverage with a tripled number of event types and 10 times the number of argument roles relative to ACE/ERE.
Figure~\ref{fig:ace-in-geneva} further depicts how ACE/ERE focus only on specific abstractions Action and Change, while \dataName{} is the most diverse with events ranging from $5$ abstract types.

\noindent \textbf{Challenging:}
The average number of mentions per event type and argument role (Table~\ref{tab:dataset-coverage}) is relatively less for \dataName{}.
Consequently, EAE models need to train from fewer examples on average which makes training more challenging.





\begin{figure}[t]
    \centering
    \includegraphics[width=0.48\textwidth]{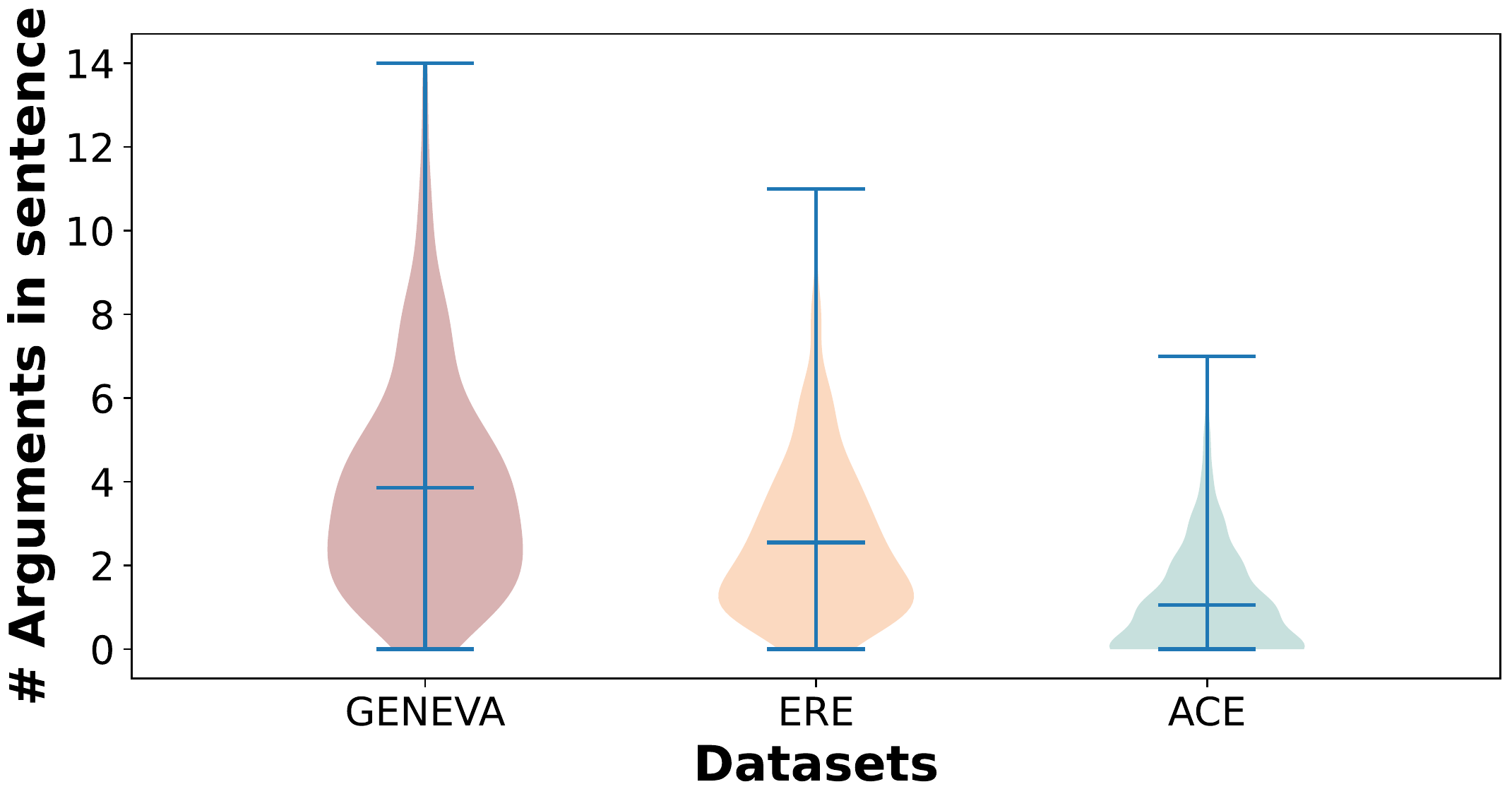}
    \caption{Violin plots for number of arguments per sentence for ACE, ERE and \dataName{} datasets.
    }
    \label{fig:argument-mention-density}
    \vspace{-3mm}
\end{figure}

\noindent \textbf{Dense:}
We plot the distribution of arguments per sentence\footnote{We remove no event mention sentences for ACE/ERE.} for ACE, ERE, and \dataName{} in Figure~\ref{fig:argument-mention-density}.
We note that \dataName{} has the highest density of $4$ argument mentions per sentence.
Both ACE and ERE have more than $70\%$ sentences with up to 2 arguments.
In contrast, \dataName{} is denser with almost $50\%$ sentences having 3 or more arguments.

\noindent \textbf{Coverage:}
Qualitatively, we show some coverage of diverse examples in Figure 9 (Appendix) and provide coverage for all events categorized by their abstraction in Figure 14 (Appendix).
We observe frequent events like Statement, Arriving, Action while Recovering, Emergency, Hindering are less-frequent events. In terms of diversity of data sources, our data comprises a mixture of news articles, Wall Street Journal articles, books, Wikipedia, and other miscellaneous sources too.

\subsection{Benchmarking Test Suites}
\label{sec:benchmark-setup}

With a focus on the generalizability evaluation of EAE models, we fabricate four benchmarking test suites clubbed into two higher-level settings:

\mypar{Limited Training Data:}
This setting mimics the realistic scenario when there are fewer annotations available for the target events and evaluates the models' ability to learn from limited training data.
We present two test suites for this setting: 
\begin{itemize}[topsep=1pt, itemsep=-3pt, leftmargin=13pt]
    \item Low resource (LR): Training data is created by \textit{randomly} sampling $n$ event mentions.\footnote{To discount the variance of the number of event mentions per sentence, we create the sampled training data such that each of them has a fixed number of $n$ event mentions.}
    We record the model performance across a spectrum from extremely low resource ($n=10$) to moderate resource ($n=1200$) settings.
    \item Few-shot (FS): Training data is curated by sampling $n$ event mentions \textit{uniformly} across all events. This sampling strategy avoids biases towards high data events and assesses the model's ability to perform well uniformly across events. We study the model performance from one-shot ($n=1$) to five-shot ($n=5$).
\end{itemize}

\mypar{Unseen Event Data:}
The second setting focuses on the scenario when there is no annotation available for the target events. This helps test models' ability to generalize to unseen events and argument roles.
We propose two test suites: 

\begin{itemize}[topsep=1pt, itemsep=-3pt, leftmargin=13pt]
    \item Zero-shot (ZS): The training data comprises the top $m$ events with most data, where $m$ varies from 1 to 10.\footnote{We sample a fixed $450$ sentences for training to remove the variance of dataset size for different $m$.}
    The remaining 105 events are used for evaluation.
    \item Cross-type Transfer (CTT):
    We curate a training dataset comprising of events of a single abstraction category (e.g. Scenario), while the test dataset comprises events of all other abstraction types. This test suite also assesses models' transfer learning strength.
\end{itemize}


Data statistics for these suites are presented in Appendix~\ref{sec:data-statistics}.
For each setup, we sample $5$ different datasets\footnote{All datasets will be released for reproducibility purpose.} and report the average model performance to account for the sampling variation.
\looseness=-1


\section{Experimental Setup}

We evaluate the generalizability of various EAE models on \dataName.
We describe these models in \S~\ref{sec:models} and the evaluation metrics in \S~\ref{sec:eval-metrics}.

\subsection{Benchmarked Models}
\label{sec:models}



Overall, we benchmark six EAE models from various representative families are described below. Implementation details are specified in Appendix~\ref{sec:implementation}.

\mypar{Classification-based models:}
These traditional works predict arguments by learning to trace the argument span using a classification objective.
We experiment with three models:
(1) \textbf{DyGIE++} \cite{wadden-etal-2019-entity}, a traditional model utilizing multi-sentence BERT encodings and span graph propagation.
(2) \textbf{OneIE} \cite{lin-etal-2020-joint}, a multi-tasking objective-based model exploiting global features for optimization.
(3) \textbf{Query\&Extract} \cite{wang-etal-2022-query} utilizing the attention mechanism to extract arguments from argument role queries.

\mypar{Question-Answering models:}
Several works formulate event extraction as a machine reading comprehension task.
We consider one such model -
(4) \textbf{BERT\_QA} \cite{du-cardie-2020-event}, a BERT-based model leveraging label semantics using a question-answering objective.
In order to scale BERT\_QA to the wide range of argument roles, we generate question queries of the form ``\textit{What is \{arg-name\}?}'' for each argument role \textit{\{arg-name\}}.
(5) \textbf{TE} \cite{lyu-etal-2021-zero}, a zero-shot transfer model that utilizes an existing pre-trained textual entailment model to automatically extract events.
Similar to BERT\_QA, we design hypothesis questions as ``\textit{What is \{arg-name\}?}'' for each argument role \textit{\{arg-name\}}.

\mypar{Generation-based models:}
Inspired by great strides in natural language generation, recent works frame EAE as a generation task using a language-modeling objective.
We consider two such models:
(6) \textbf{TANL} \cite{Paolini21tacl}, a multi-task language generation model which treats EAE as a translation task.
(7) \textbf{DEGREE} \cite{hsu2021degree}, an encoder-decoder framework that extracts event arguments using natural language input prompts.

\noindent \textit{Automating DEGREE}:
DEGREE requires human effort for manually creating natural language prompts and thus, can not be directly deployed for the large set of event types in \dataName.
In our work, we undertake efforts to scale up DEGREE by proposing a set of automated refinements.
The first refinement automates the event type description as ``\textit{The event type is \{event-type\}}" where \textit{\{event-type\}} is the input event type.
The second refinement automates the event template generation by splitting each argument into a separate self-referencing mini-template ``\textit{The \{arg-name\} is some \{arg-name\}}" where \textit{\{arg-name\}} is the argument role.
The final event-agnostic template is a simple concatenation of these mini-templates.
We provide an illustration and ablation of these automated refinements for DEGREE in Appendix~\ref{sec:autodegree-ablation}.

\begin{figure}[t]
    \centering
    \includegraphics[width=0.48\textwidth]{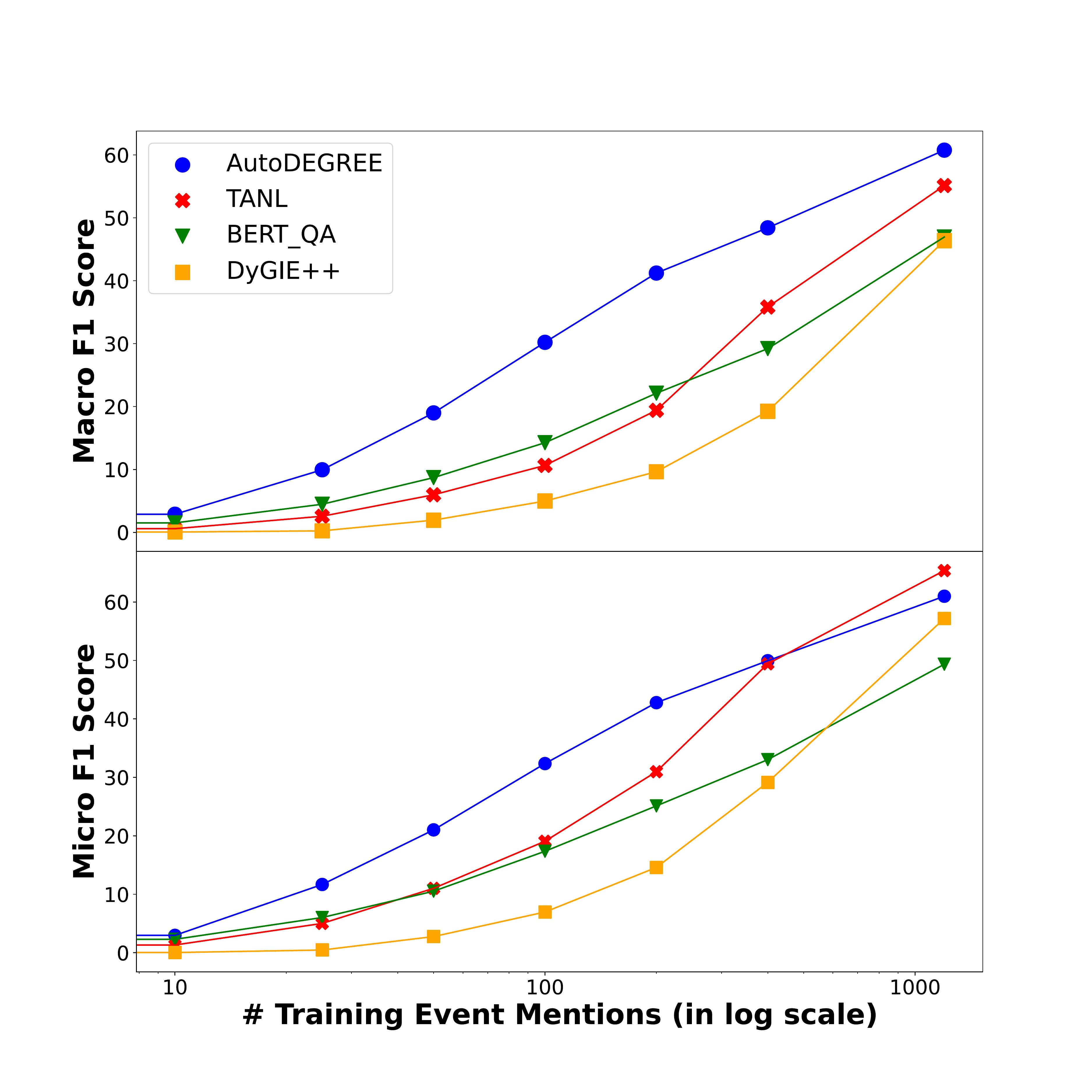}
    \caption{
    \looseness=-1
    Model performance in macro F1 (top) and micro F1 (bottom) scores against the number of training event mentions (log-scale) for the low resource suite. Each datapoint is an average of 5 runs.
    }
    \label{fig:lr_results}
    \vspace{-1mm}
\end{figure}

\begin{figure}[t]
    \centering
    \includegraphics[width=0.48\textwidth]{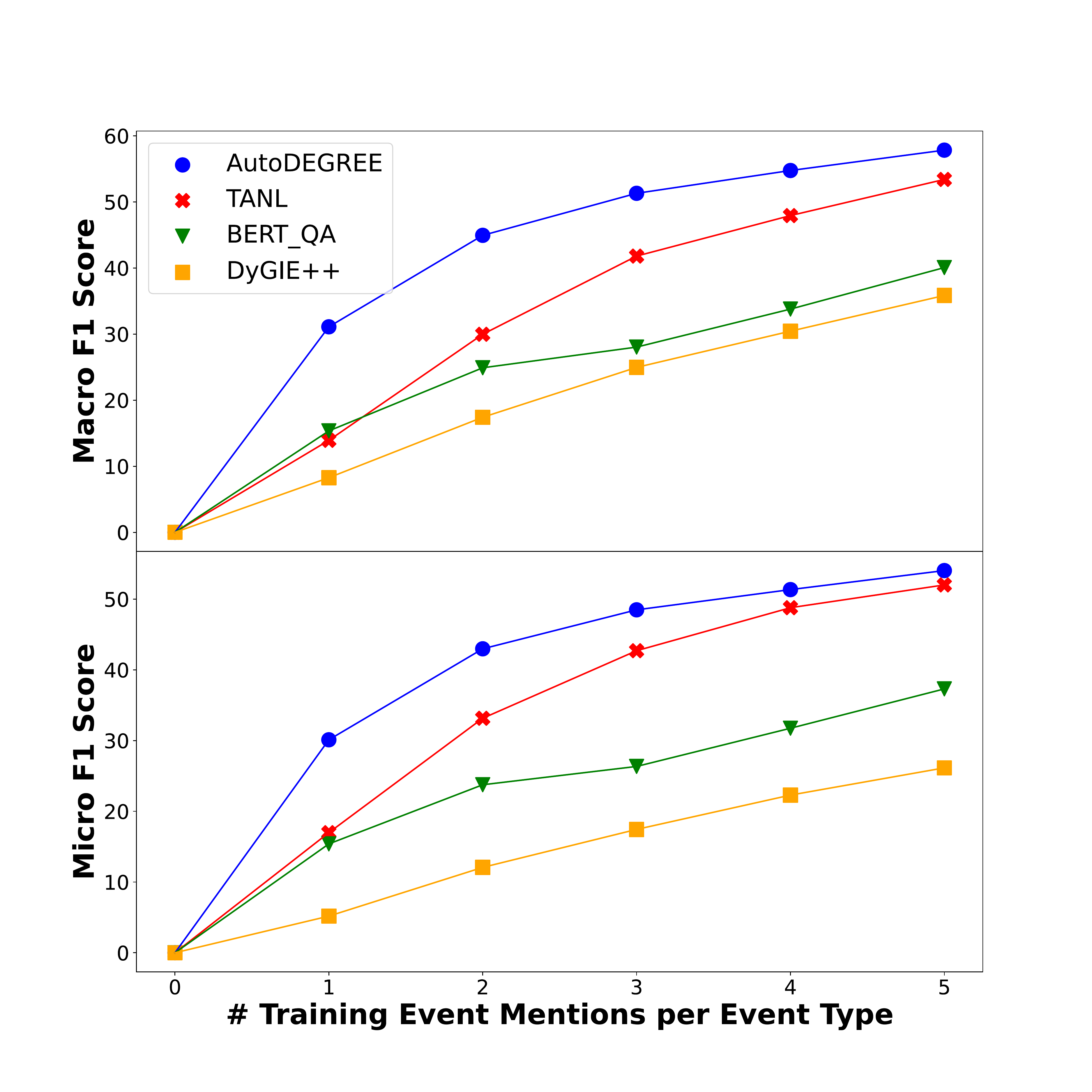}
    \caption{Model performance in macro F1 (top) and micro F1 (bottom) scores against the number of training event mentions per event for the few-shot suite. Each datapoint is an average of 5 runs.
    }
    \label{fig:fs_results}
    \vspace{-1mm}
\end{figure}

\subsection{Evaluation Metrics}
\label{sec:eval-metrics}

Following the traditional evaluation for EAE tasks, we report the \textbf{micro F1} scores for argument classification.
To encourage better generalization across a wide range of events, we also use \textbf{macro F1} score that reports the average of F1 scores for each event type.
For the limited data test suites, we record a model performance curve, wherein we plot the F1 scores against the number of training instances.


\section{Results}
\label{sec:results}

Following \S~\ref{sec:benchmark-setup}, we organize the main experimental results into limited training data and unseen event data settings.
When trained on complete training data, we observe that 
OneIE and Query\&Extract models achieve poor micro F1 scores of just $30.03$ and $40.41$ while all other models achieve F1 scores above $55$.
This can be attributed to the inability of their model designs to effectively handle overlapping arguments.\footnote{One key attribute of \dataName{} is that arguments overlap with each other quite frequently in a sentence.}
Due to their inferior performance, we do not include OneIE and Query\&Extract in the benchmarking results.
We present the full results in Appendix~\ref{sec:all-results}.

\subsection{Limited Training Data}



Limited training data setting comprises of the low resource and the few-shot test suites.
We present the model benchmarking results in terms of macro and micro F1 scores for the low resource test suite in Figure~\ref{fig:lr_results} and for the few-shot test suite in Figure~\ref{fig:fs_results} respectively.
We observe that \modelName{} outperforms all other models
for both the test suites and shows superior generalizability.
In general, we observe that generation-based models show better generalization while on the other hand, traditional classification-based approaches show poor generalizability.
This underlines the importance of using label semantics for better generalizability.
We also detect a stark drop from micro to macro F1 scores for TANL and DyGIE++ in the low resource test suite.
This indicates that these models are more easily biased toward high data events and do not generalize well uniformly across all events.

\subsection{Unseen Event Data}

This data setting includes the zero-shot and the cross-type transfer test suites.
We collate the results in terms of micro F1 scores
for both the test suites in Table~\ref{tab:no-train-results}.
Models like DyGIE++ and TANL cannot support unseen events or argument roles
and thus, we do not include these models in the experiments for these test suites.
TE cannot be trained on additional EAE data, and hence we only report the pure zero-shot performance of this model.

From Table~\ref{tab:no-train-results}, we observe that \modelName{} achieves the best scores across both test suites outperforming BERT\_QA by a significant margin of almost $13$-$15\%$ F1 points.
Although TE is not comparable as it's a pure zero-shot model (without training on any data), it's performance is relatively super low in both settings.
Thus, \modelName{} shows superior transferability to unseen event types and argument roles.

\begin{table}[t]
    \centering
    \small
    \begin{tabular}{l|rrr|r}
        \toprule
        \textbf{Model} & \textbf{ZS-1} & \textbf{ZS-5} & \textbf{ZS-10} & \textbf{CTT} \\
        \midrule
        TE$^*$ & 7.54 & 7.54 & 7.54 & 6.39 \\
        BERT\_QA & 5.05 & 21.53 & 24.24 & 11.17 \\
        \modelName & \textbf{24.06} & \textbf{34.68} & \textbf{39.43} & \textbf{27.9} \\
        \bottomrule
    \end{tabular}
    \caption{Model performance in micro F1 scores for the zero-shot (ZS) and cross-type transfer (CTT) test suites. ZS-1, ZS-5, and ZS-10 indicate 1, 5, and 10 event types for training respectively. Each datapoint is an average of 5 runs. $^*$Not directly comparable as TE doesn't train on any data.
    }
    \label{tab:no-train-results}
\end{table}

\section{Analysis}

In this section, we provide analyses highlighting the various new challenges introduced by \dataName{}.
We discuss the performance of large language models, the introduction of non-entity argument roles, and model performance including Time and Place argument roles.

\subsection{Large Language Model Performance}

\begin{figure}[t]
    \centering
    \includegraphics[width=0.48\textwidth]{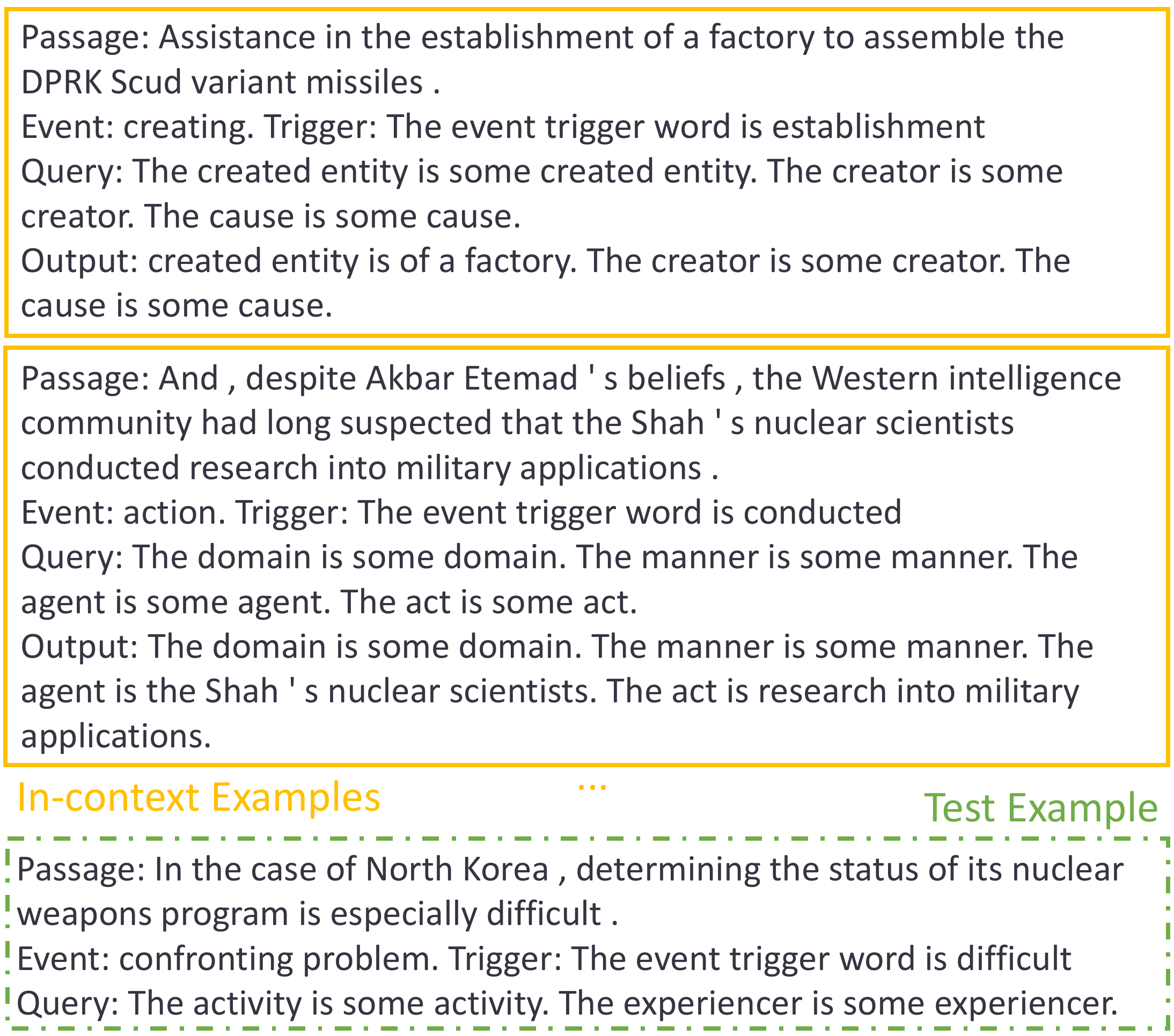}
    \caption{Illustration of the prompt used for evaluating GPT3.5-turbo. We provide 5 in-context examples (solid yellow) and provide the test example (dashed green).}
    \label{fig:llm-prompt}
    \vspace{-4mm}
\end{figure}

Recently, there has been an advent of Generative AI in the form of Large Language Models (LLMs) like GPT-3 \cite{brown2020language}, GPT-4, PaLM \cite{chowdhery2022palm}, Code4Struct \cite{wang2022code4struct}, and many more.
We evaluate one of these models GPT3.5-turbo on the task of EAE on the zero-shot test suite of \dataName \footnote{Since we can't fine-tune LLMs on known event types, this is not the most fair comparison, but the closest one possible.}.
More specifically, we provide 5 in-context examples from top-10 events and evaluate test examples from the remaining 105 events.
Our GPT-prompt template follows the DEGREE template wherein model replaces placeholders with arguments if present, else copies the original teample. An illustration is provided in Figure~\ref{fig:llm-prompt}.

Despite the strong generation capability, GPT3.5-turbo achieves a mere $\textbf{22.73}$ F1 score while DEGREE achieves $\textbf{24.06}$ and $\textbf{39.43}$ F1 scores in the ZS-1 and ZS-10 test suites respectively. Although these scores aren’t directly comparable, it shows how GENEVA is quite challenging for LLMs in the zero-shot/few-shot setting.
\looseness=-1

\subsection{New Challenge of Non-entity Roles}

\begin{table}[t]
    \centering
    \small
    \begin{tabular}{l|cc|cc}
        \toprule
         & \multicolumn{2}{c|}{\textbf{LR-400}} & \multicolumn{2}{c}{\textbf{ZS-10}} \\
         & GENEVA & ACE & GENEVA & ACE \\
         \midrule
         BERT\_QA & $33$ & - & $24.2$ & $46.7^*$ \\
         DEGREE & $49.9$ & $57.3^*$ & $39.4$ & $53.3^*$ \\
         \bottomrule
    \end{tabular}
    \caption{Model performance in micro F1 for BERT\_QA and DEGREE for low resource with 400 training mentions (LR-400) and zero-shot with 10 training events (ZS-10) test suites across \dataName{} and ACE. $^*$Reported from \citet{hsu-etal-2022-degree}.}
    \label{tab:ace-geneva-ablation}
\end{table}

In Table~\ref{tab:ace-geneva-ablation}, we show the model performances of BERT\_QA and DEGREE on \dataName{} and ACE under similar benchmarking setups.
We note how both models exhibit relatively poor performance on \dataName{} (especially the zero-shot test suite).
To investigate this phenomenon, we break down the model performance based on entity and non-entity argument roles and show this analysis in Table~\ref{tab:entity-analysis}.
This ablation reveals a stark drop of $10$-$14\%$ F1 points across all models when predicting non-entity arguments relative to entity-based arguments.
This trend is observed consistently across all different test suites as well.
We can attribute this difference in model performance to non-entity arguments being more abstract and having longer spans, in turn, being more challenging to predict accurately.
Thus, owing to a significant $37\%$ non-entity argument roles, \dataName{} poses a new and interesting challenge for generalization in EAE.



\subsection{\dataName{} with Time and Place}

In the original \dataName{} dataset, we filtered super generic argument roles, but some of these roles like Time and Place are key for several downstream tasks.
We include Time and Place arguments in \dataName\footnote{We release this data for future development} and provide results of the models on the full dataset in Table~\ref{tab:results-time-place}.
Compared to original \dataName{} results in the same setting, we observe a slight dip in the model performance owing to the addition of extra arguments.
Overall, the trend is similar where TANL performs the best and we observe better generalization in terms of macro F1 performance.

\begin{table}[t]
    \centering
    \small
    \begin{tabular}{l|rrr}
        \toprule
         & \textbf{Entity} & \textbf{Non-entity} & \textbf{$\Delta$} \\
        \midrule
        DEGREE & $54.46$ & $39.89$ & $14.57$ \\
        TANL & $52.54$ & $42.4$ & $10.14$ \\
        BERT\_QA & $36.71$ & $24.86$ & $11.85$ \\
        \bottomrule
    \end{tabular}
    \caption{Breakdown of micro F1 scores into the entity and non-entity arguments for DEGREE, TANL, and BERT\_QA models on the low resource setting with $400$ training mentions. $\Delta$ denotes the difference.}
    \label{tab:entity-analysis}
\end{table}

\subsection{Discussion}

Overall, our generalizability benchmarking reveals various insights.
First, generation-based models like DEGREE exhibit strong generalizability and establish a benchmark on our dataset.
Second, macro score evaluation reveals how models like TANL and DyGIE++ can be easily biased toward high-data events.
Finally, we show how \dataName{} poses a new challenge in the form of non-entity arguments, encouraging further research for improving generalization in EAE.

\begin{table}[t]
    \centering
    \small
    \begin{tabular}{l|rr}
        \toprule
        \textbf{Model} & \textbf{Micro F1} & \textbf{Macro F1} \\
        \midrule
        BERT\_QA & $52.97$ & $50.16$ \\
        DyGIE++ & $65.03$ & $54.85$ \\
        TANL & $\textbf{71.17}$ & $\textbf{65.18}$ \\
        \modelName & $59.74$ & $59.20$ \\
        \bottomrule
    \end{tabular}
    \caption{Model performance in micro F1 and macro F1 scores for the full \dataName{} dataset with Time and Place arguments.}
    \label{tab:results-time-place}
\end{table}

\section{Conclusion and Future Work}


In our work, we exploit the shared relations between SRL and EAE to create a new large and diverse event argument ontology spanning $115$ event types and $220$ argument roles.
This vast ontology can be used to create larger and more comprehensive resources for event extraction.
We utilize this ontology to build a new generalizability benchmarking dataset \dataName{} comprising four distinct test suites and benchmark EAE models from various families.
Our results inspire further research of generative models for EAE to improve generalization.
Finally, we show that \dataName{} poses new challenges and anticipate future generalizability benchmarking efforts on our dataset.


\section*{Acknowledgements}

We would like to thank Hritik Bansal, Di Wu, Sidi Lu, Derek Ma, Anh Mac, and Zhiyu Xie for their valuable insights, experimental setups, paper reviews, and constructive comments. 
We thank the anonymous reviewers for their feedback. 
This work was partially supported by NSF 2200274, AFOSR MURI via Grant \#FA9550- 22-1-0380, Defense Advanced Research Project Agency (DARPA) grant \#HR00112290103/HR0011260656, and a Cisco Sponsored Research Award.

\section*{Limitations}

We would like to highlight a few limitations of our work.
First, we would like to point out that \dataName{} is designed to evaluate the generalizability of EAE models.
Although the dataset contains event type and event trigger annotations, it can only be viewed as a partially-annotated dataset if end-to-end event extraction is considered.
Second, \dataName{} is derived from an existing dataset FrameNet.
Despite human validation efforts, there is no guarantee that all possible events in the sentence are exhaustively annotated.

\section*{Ethical Consideration}

We would like to list a few ethical considerations for our work.
First, \dataName{} is derived from FrameNet which comprises of annotated sentences from various news articles.
Many of these news articles cover various political issues which might be biased and sensitive to specific demographic groups.
We encourage careful consideration for utilizing this data for training models for real-world applications.

\bibliography{anthology,custom}
\bibliographystyle{acl_natbib}

\clearpage

\appendix




\section{Additional Analysis of \dataName{}}

\subsection{Event Type Distribution for \dataName}

\begin{figure}[h]
    \centering
    \includegraphics[width=0.48\textwidth]{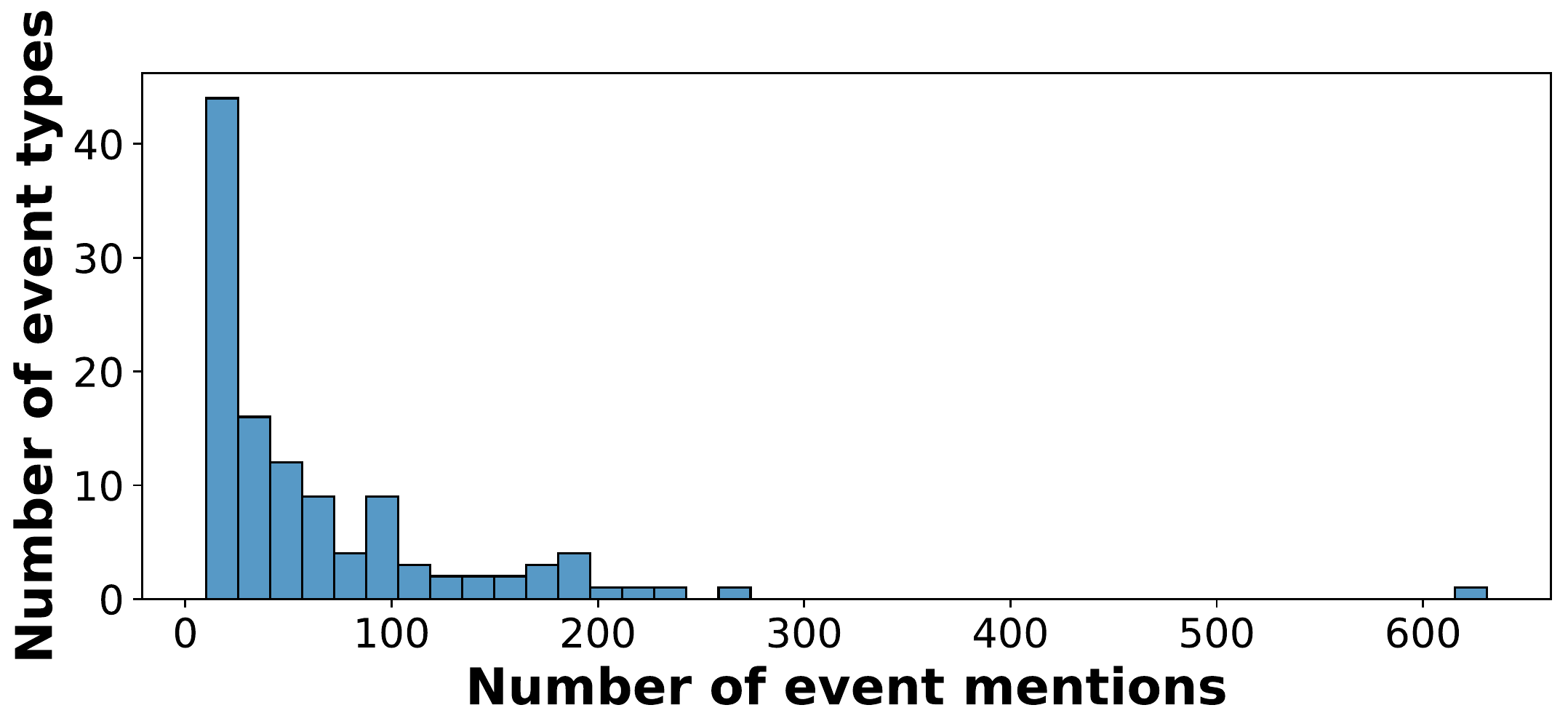}
    \caption{Distribution of event types by the number of event mentions in \dataName.}
    \label{fig:event-data-distribution}
\end{figure}

We show the distribution of event mentions per event type for \dataName{} in Figure~\ref{fig:event-data-distribution}.
We observe a highly skewed distribution with $44$ event types having less than 25 event mentions.
Furthermore, $93$ event types have less than 100 event mentions.
We believe that this resembles a more practical scenario where there is a wide range of events with limited event mentions while a few events have a large number of mentions.

\subsection{Data Statistics
for different benchmarking test suites}
\label{sec:data-statistics}

We present the data statistics for the various test suites in Table~\ref{tab:data-split}.
For the training set of the low resource and few-shot test suites (indicated by $*$ in Table~\ref{tab:data-split}), we sample a smaller training set (as discussed in Section~\ref{sec:benchmark-setup}).
For the zero-shot setup, the top 10 event types contribute to a large pool of $1,889$ sentences.
For the test suites, a fixed number of 450 and 115 sentences are sampled for the training and the development set (indicated by $+$ in Table~\ref{tab:data-split}) from this larger pool of data.

\begin{table}[h]
    \centering
    \small
    \begin{tabular}{lccc}
        \toprule
        & \textbf{LR}/\textbf{FS} & \textbf{ZS} & \textbf{CTT} \\
        \midrule
        \# Train Sentences & $1,967^*$ & $450^+$ & $268$ \\
        \# Dev Sentences & $778$ & $115^+$ & $66$ \\
        \# Test Sentences & $928$ & $1,784$ & $3,339$ \\
        \bottomrule
    \end{tabular}
    \caption{Data statistics of the number of test sentences for the different benchmarking test suites. Here, LR: Low Resource, FS: Few-shot, ZS: Zero-shot, CTT: Cross-Type Transfer. $*$ and $+$ indicate that certain sampling is done for creating these datasets. More details are provided in the text.
    }
    \label{tab:data-split}
\end{table}


\begin{figure*}[t]
    \centering
    \includegraphics[width=0.96\textwidth]{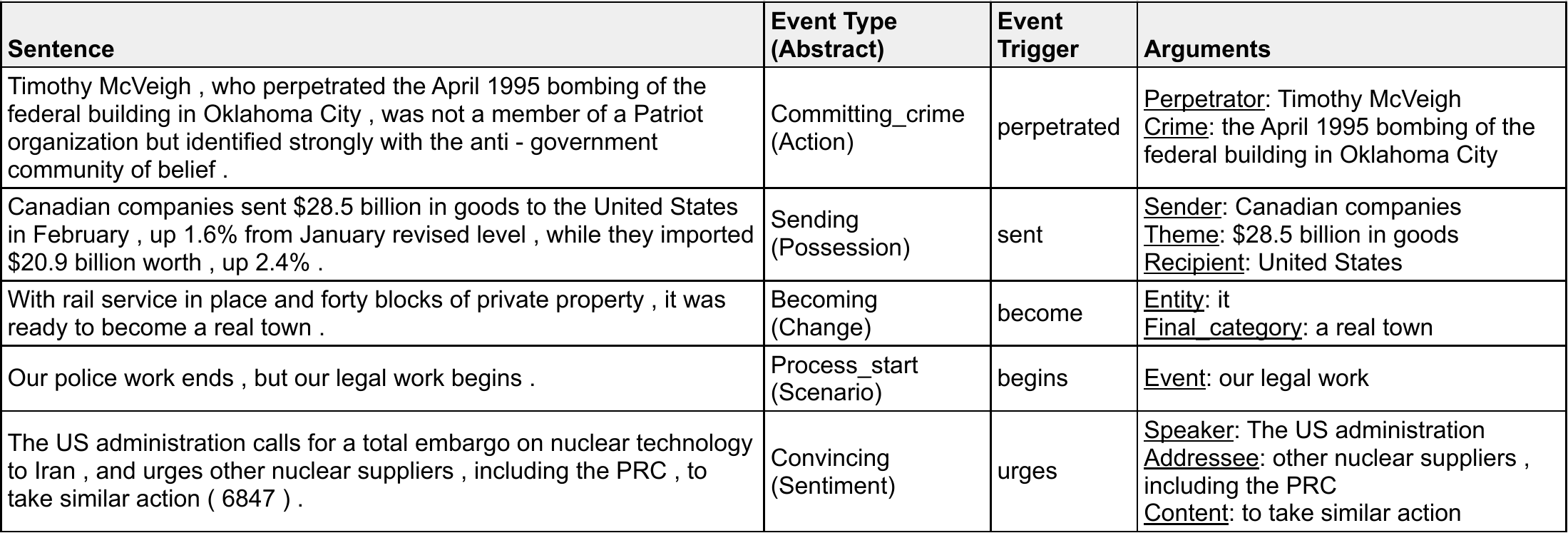}
    \caption{Illustration of example annotations from the \dataName{} dataset for various different abstract types.}
    \label{fig:dataset-examples}
\end{figure*}

\subsection{Event Ontology Organization}
\label{sec:event-organization}

The broad set of event types in \dataName{} can be organized into a hierarchical structure of abstract event types.
Adhering to the hierarchical tree structure introduced in MAVEN, we show the corresponding organization for event types in \dataName{} in Figure~\ref{fig:event-organization}.
The organization mainly assumes five abstract event categories - Action, Change, Scenario, Sentiment, and Possession.
The most populous abstract type is Action with a total of 53 events, while Scenario abstraction has the lowest number of 9 events.

We also study the distribution of event mentions per event type in Figure~\ref{fig:event-organization} where the bar heights are indicative of the number of event mentions for the corresponding event type (heights in log-scale).
We observe that the most populous event is \textit{Statement} which falls under the Action abstraction.
On the other hand, the least populous event is \textit{Recovering} which belongs to the Change abstraction.



\dataName{} comprises of a diverse set of 115 event types and it naturally shares some of these with the ACE dataset.
In Figure~\ref{fig:event-organization}, we show the extent of the overlap of the mapped ACE events in the \dataName{} event schema (text labels colored in red).\footnote{We only show the events that could be directly mapped from ACE to GENEVA. Note that this overlap is not exhaustively complete. Furthermore, the mapping can be many-to-one and one-to-many in nature.}
We can observe that although there is some overlap between the datasets, \dataName{} brings in a vast pool of new event types.
Furthermore, most of the overlap is for the Possession and Action abstraction types.


\subsection{Dataset Examples}

We provide some examples of annotated sentences from the \dataName{} dataset in Figure~\ref{fig:dataset-examples}.
We indicate the abstract event type in braces and cover an example from each abstraction.

\section{Automated Refinements for DEGREE}
\label{sec:autodegree-ablation}

\subsection{DEGREE}

\begin{figure}[t]
    \centering
    \includegraphics[width=0.45\textwidth]{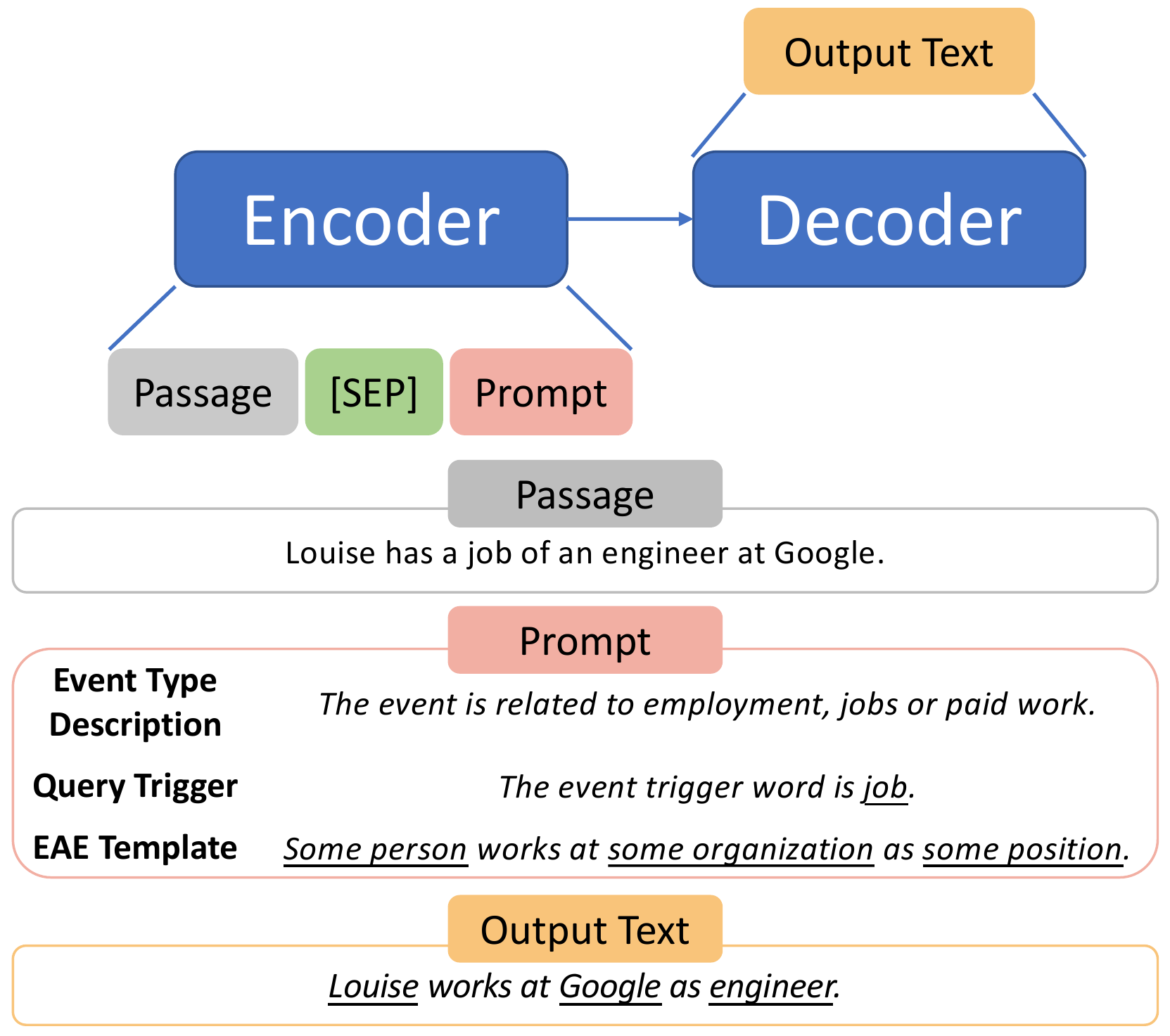}
    \caption{Model architecture of \degree{} (top half) and an illustration of a manually created prompt for the event type \textit{Employment} (bottom half).
    }
    \label{fig:degree-eae}
    \vspace{-0.7em}
\end{figure}

DEGREE is an encoder-decoder based generative model which 
utilizes natural language templates as part of input prompts.
The input prompt comprises of three components - (1) \textit{Event Type Description} which provides a definition of the given event type, (2) \textit{Query Trigger} which indicates the trigger word for the event mention, and (3) \textit{EAE Template} which is a natural sentence combining the different argument roles of the event.
We illustrate DEGREE along with an example of its input prompt design in Figure~\ref{fig:degree-eae}.

Despite the superior performance of \degree{} in the low-data setting, it can not be directly deployed on \dataName{}.
This is because \degree{} requires manual human effort for the creation of input prompts for each event type and argument role and can't be scaled to the wide set of events in \dataName.
Thus, there is a need to automate the manual human effort to scale up \degree.

\begin{figure}[t]
    \centering
    \includegraphics[width=0.48\textwidth]{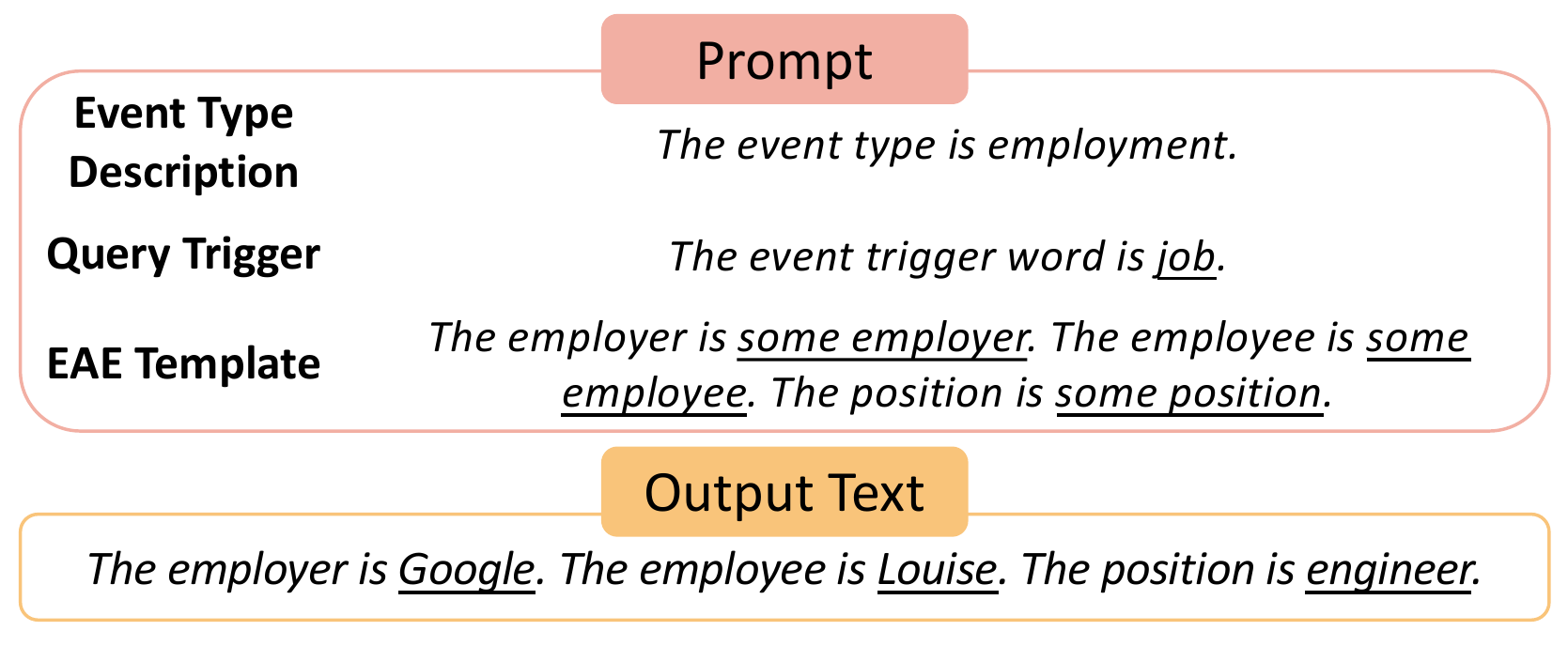}
    \caption{An illustration of an automatically generated prompt by \modelName{} for the event type \textit{Employment}.}
    \label{fig:automated-prompt}
    \vspace{-0.7em}
\end{figure}

\subsection{Automated Refinements}
\label{sec:dam}

\degree{} requires human effort for two input prompt components - (1) Event Type Description and (2) EAE Template.
We describe the automated refinements in \modelName{} for these components below.

\paragraph{Automating Event Type Description}
Event type description is a natural language sentence describing the event type.
In order to automate this component, we propose a simple heuristic that creates a simple natural language sentence mentioning the event type - ``\textit{The event type is \{event-type\}}.", as illustrated in Figure~\ref{fig:automated-prompt}.

\paragraph{Automating EAE Template}
EAE template generation in \degree{} can be split into two subtasks, which we discuss in detail below.

\textit{Argument Role Mapping:}
This subtask maps each argument role to a natural language placeholder phrase based on the characteristics of the argument role.
For example, the argument role \textit{Employer} is mapped to ``\textit{some organization}'' in Figure~\ref{fig:degree-eae}.
For automating this mapping process, we propose a simple refinement of self-mapping, which maps each argument role to a self-referencing placeholder phrase ``\textit{some \{arg-name\}}'', where \textit{\{arg-name\}} is the argument role itself.
For example, the argument role \textit{Employer} would be mapped to ``\textit{some employer}''.
We illustrate an example of this heuristic in Figure~\ref{fig:automated-prompt}.

\textit{Template Generation:}
The second subtask requires generating a natural sentence(s) using the argument role-mapped placeholder phrases (as shown in Figure~\ref{fig:degree-eae}).
To automate this subtask, we create an event-agnostic template composed of argument role-specific sentences.
For each argument role in the event, we generate a sentence of the form ``\textit{The \{arg-name\} is \{arg-map\}.}'' where \textit{\{arg-name\}} and \textit{\{arg-map\}} is the argument role and its mapped placeholder phrase respectively.
For example, the sentence for argument role \textit{Employer} with self-mapping would be "\textit{The employer is some employer.}".
The final event-agnostic template is a simple concatenation of all the argument role sentences.
We provide an illustration of the event-agnostic template in Figure~\ref{fig:automated-prompt}.

\subsection{Ablation Study}

\begin{table}[t]
    \centering
    \small
    \begin{tabular}{l|rr}
        \toprule
         & \textbf{Original} & \textbf{Automated} \\
         & \textbf{DEGREE} & \textbf{DEGREE} \\
        \midrule
        ACE Dataset & 73.5 & 72.7 \\
        \bottomrule
    \end{tabular}
    \caption{Model Performance in terms of F1 score for \degree{} and \modelName{} on the ACE dataset.}
    \label{tab:ablations-ace}
\end{table}

In our work, we introduce automated refinements for scaling DEGREE for \dataName.
We provide an ablation study for these automated refinements (Automated DEGREE) on the ACE dataset in Table~\ref{tab:ablations-ace}.
We observe that the automated DEGREE almost at-par with DEGREE with  a minor difference of only $0.8\%$ F1 points.




\section{Impact of Pre-training}
\label{sec:pre-training}

\begin{figure}[t]
    \centering
    \includegraphics[width=0.48\textwidth]{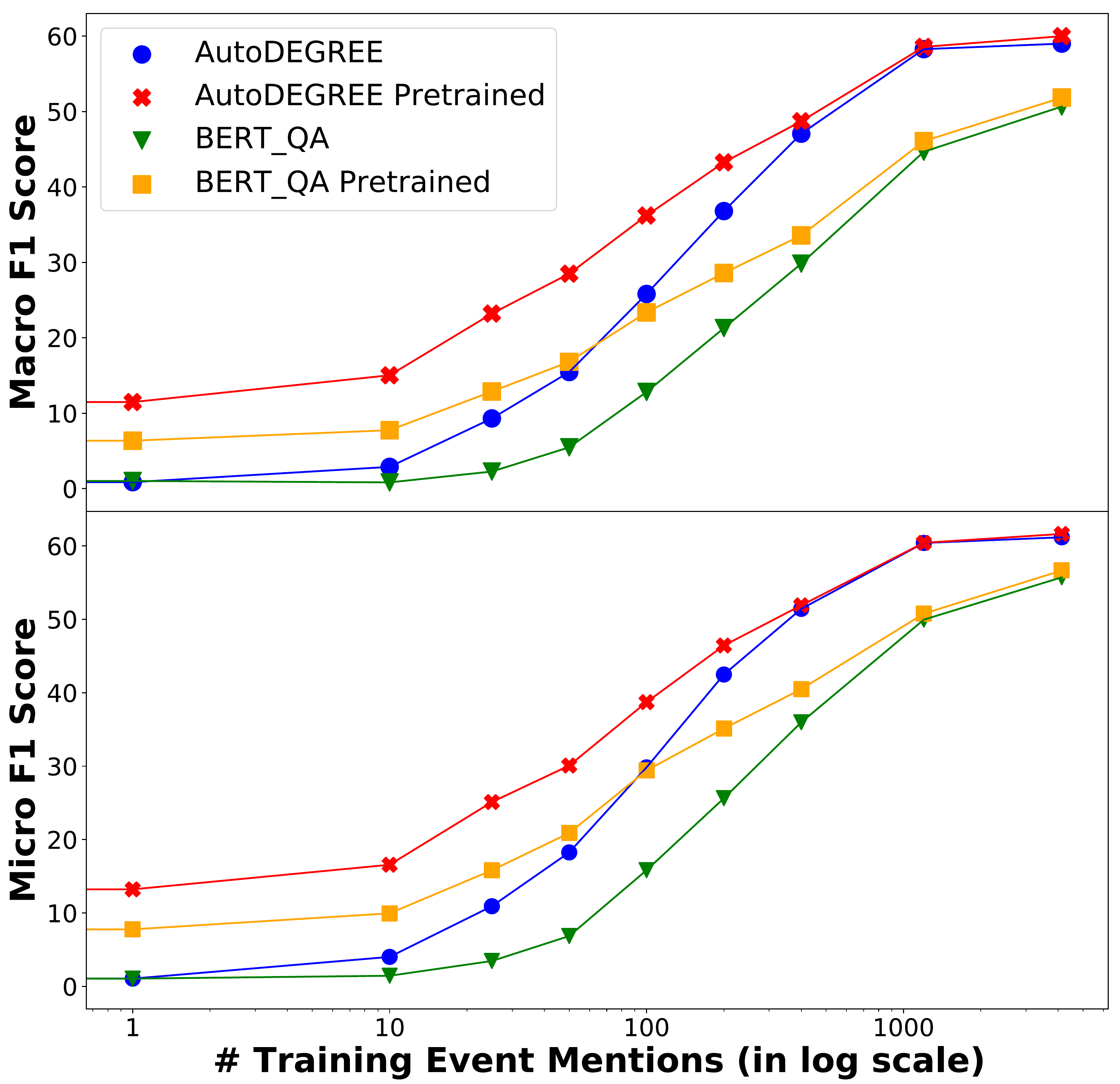}
    \caption{
    Model performance in macro F1 (top) and micro F1 (bottom) scores against the number of training event mentions (log-scale) for the low resource suite. Here we majorly compare the impact of pre-training on the model performance.
    }
    \label{fig:lr_pretraining_results}
\end{figure}

In this section, we explore the impact of pre-training models on the generalizability evaluation.
We consider DEGREE and BERT\_QA, pre-train them on the ACE dataset and show the model performance on low resource test suite in 
Figure~\ref{fig:lr_pretraining_results}.

We observe that pre-training helps model performance by $5$-$10\%$ F1 points, and naturally in the low-data regime.
But the gains diminish and are almost negligible as the number of training event mentions increases.
In terms of zero-shot performance of the pre-trained models, DEGREE achieves a micro F1 score of $12.83\%$ and BERT\_QA achieves a score of $6.82\%$ respectively.
Poor zero-shot performance and diminishing performance gains indicate that \dataName{} is distributionally distinct from ACE, which makes it challenging to achieve good model performance on \dataName{} merely via transfer learning.

\section{Case Study: Is ACE diverse enough?}
\label{sec:case-study-diversity}

We conduct a case study to analyze how the limited diversity of ACE can affect the generalizability of EAE models.
We compare the performance of two models with different initializations - (1) \modelName{} pre-trained on the ACE dataset and (2) \modelName{} with no pre-training - on the zero-shot with 10 event types benchmarking setup.
We dissect the F1 scores into different abstract event types and show the results in Table~\ref{tab:case-study-diversity}.

\begin{table}[t]
    \centering
    \small
    \begin{tabular}{l|rrr}
        \toprule
        \textbf{Abstract} & \textbf{Scratch} & \textbf{Pre-Trained} & $\Delta$ \\
        \textbf{Event Type} & \textbf{Model} & \textbf{Model} \\
        \midrule
        Action & $35.48$ & $38.93$ & $3.45$ \\
        Possession & $45.65$ & $50.63$ & $4.98$ \\
        Change & $38.5$ & $43.4$ & $4.9$ \\
        Sentiment & $49.37$ & $51.55$ & $2.18$ \\
        Scenario & $30.87$ & $34.59$ & $3.72$ \\
        \bottomrule
    \end{tabular}
    \caption{Model Performance in micro F1 on zero-shot with 10 event types split by abstract event types for (1) \modelName{} with no pre-training (Scratch Model), and (2) Pre-Trained \modelName{} on ACE (Pre-Trained Model). $\Delta$: model performance difference.}
    \label{tab:case-study-diversity}
\end{table}

We observe that pre-training yields major improvements for the abstractions of Action, Possession, and Change - which are well-represented in ACE.
On the other hand, we observe lower performance improvement for the abstractions of Sentiment and Scenario - which are not represented in ACE.
This trend clearly shows that the lack of diversity in ACE restricts the models' ability to generalize well to out-of-domain event types.
We also highlight the significance of \dataName{} as its diverse evaluation setup helps analyze these trends.

\begin{figure*}[t]
    \centering
    \includegraphics[width=0.98\textwidth]{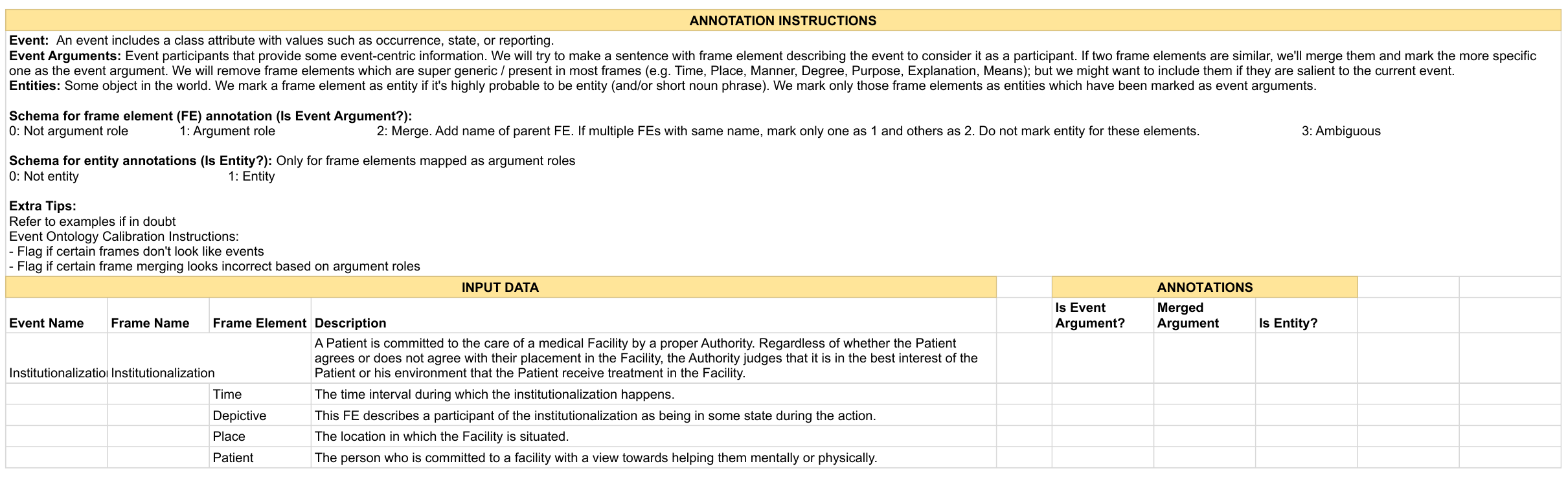}
    \caption{Figure illustrating the annotation process for EAE ontology creation. At the top, we present the annotation instructions. In the second bottom half, we show how the input data is presented along with fields for annotations.}
    \label{fig:ontology-annotation}
\end{figure*}

\section{Human expert annotation for EAE ontology creation}
\label{sec:ontology-annotation-appendix}

Figure~\ref{fig:ontology-annotation} present the annotation instructions and example input data for the human expert annotation process used for event argument ontology creation.

\begin{table*}[ht]
    \centering
    \small
    \begin{tabular}{lp{10cm}l}
        \toprule
         & \textbf{Sentence} & \textbf{Event Trigger} \\
        \midrule
        \textbf{Primary} & Both villages offer good waterfront restaurants with homestyle Chinese food, principally seafood fresh from the tank. & offer \\
        \textbf{Candidate 1} & It gives an overview of Macau's history and its daily life and traditions. & gives \\
        \textbf{Candidate 2} & He should do more to reduce tax rates on wealth and income, in recognition of the fact that those cuts yield higher, not lower, revenues. & revenues\\
        \bottomrule
    \end{tabular}
    \caption{Example for the human validation setup for ontology quality assessment.}
    \label{tab:ontology-quality}
\end{table*}

\begin{table*}[ht]
    \centering
    \small
    \begin{tabular}{p{6cm}|l|l|p{2.8cm}|p{2.5cm}}
        \toprule
        \textbf{Sentence} & \textbf{Event} & \textbf{Event} & \textbf{Annotated} & \textbf{Unannotated} \\
        & & \textbf{Trigger} & \textbf{Arguments} & \textbf{Arguments} \\
        \midrule
        The attackers were environmental terrorists upset about a new industry coming to town . & Attack & attackers & \underline{Assailant}: environmental terrorists & Means, Victim, Weapon \\ \hline
        United States Helps Uzbekistan Secure Dangerous Nuclear Materials : Energy agency announces completion of secret uranium transfer back to Russia & Assistance & Helps & \underline{Helper}: United States \underline{Goal}: Secure Dangerous Nuclear Materials & Benefited\_party, Focal\_entity, Means \\
        \bottomrule
    \end{tabular}
    \caption{Examples for the human validation setup for annotation comprehensiveness assessment.}
    \label{tab:annotation-comprehensive}
\end{table*}

\section{Human validation for \dataName{}}
\label{sec:human-validation-setup}

We provide an example of the annotation setup used for the \textit{Ontology Quality Assessment} as part of \dataName{} validation process in Table~\ref{tab:ontology-quality}.
Similarly, we provide the annotation setup and some examples for the \textit{Annotation Comprehensiveness Assessment} in Table~\ref{tab:annotation-comprehensive}.

\section{Implementation Details}
\label{sec:implementation}

In this section, we provide details about the experimental setups and training details for various EAE models we mentioned in our work.

\subsection{\modelName{}}

We closely follow the training setup by \degree{} for training the \modelName{} models.
We run experiments for \modelName{} on a NVIDIA GeForce RTX 2080 Ti machine with support for 8 GPUs.
We present the complete range of hyperparameter details in Table~\ref{tab:hyper-autodegree}.
We deploy early stopping criteria for stopping the model training.

\begin{table}[ht]
    \centering
    \begin{tabular}{lr}
        \toprule
        PLM & BART-Large \\
        Training Batch Size & $6$ \\
        Eval Batch Size & $12$ \\
        Learning Rate & $1 \times 10^{-5}$ \\
        Weight Decay & $1 \times 10^{-5}$ \\
        \# Warmup Epochs & $5$ \\
        Gradient Clipping & $5$ \\
        Max Training Epochs & $50$ \\
        \# Accumulation Steps & $1$ \\
        Beam Size & $1$ \\
        Max Sequence Length & $200$ \\
        Max Output Length & $150$ \\
        \bottomrule
    \end{tabular}
    \caption{Hyperparameter details for \modelName{} model.}
    \label{tab:hyper-autodegree}
\end{table}

\subsection{BERT\_QA}

We mostly follow the original experimental setup and hyperparameters as described in \citet{du-cardie-2020-event}.
We use \textsc{Bert-Large} instead of the original \textsc{Bert-Base} to ensure that the PLMs are of comparable sizes for \modelName{} and BERT\_QA.
We run experiments for this model on a NVIDIA A100-SXM4-40GB machine with support for 4 GPUs.
A more comprensive list of hyperparameters is provided in Table~\ref{tab:hyper-bert-qa}.

\begin{table}[h]
    \centering
    \begin{tabular}{lr}
        \toprule
        PLM & BERT-Large \\
        Training Batch Size & $12$ \\
        Eval Batch Size & $8$ \\
        Learning Rate & $1 \times 10^{-5}$ \\
        \# Training Epochs & $50$ \\
        \# Evaluations per Epoch & $5$ \\
        Max Sequence Length & $300$ \\
        Max Answer Length & $50$ \\
        N-Best Size & $20$ \\
        \bottomrule
    \end{tabular}
    \caption{Hyperparameter details for BERT\_QA model.}
    \label{tab:hyper-bert-qa}
\end{table}

\subsection{TANL}

We report the hyperparameter settings for the TANL experiments in Table~\ref{tab:hyper-tanl}.
We make optimization changes in the provided source code of TANL to include multiple triggers in a single sentence.
Experiments for TANL were run on a NVIDIA GeForce RTX 2080 Ti machine with support for 8 GPUs.

\begin{table}[t]
    \centering
    \begin{tabular}{lr}
        \toprule
        PLM & T5-Base \\
        Training Batch Size & $8$ \\
        Eval Batch Size & $12$ \\
        Learning Rate & $5 \times 10^{-4}$ \\
        \# Training Epochs & $20$ \\
        Evaluation per \# Steps & $100$ \\
        Max Sequence Length & $256$ \\
        \# Beams & 8 \\
        \bottomrule
    \end{tabular}
    \caption{Hyperparameter details for TANL model.}
    \label{tab:hyper-tanl}
\end{table}

\subsection{DyGIE++}

We report the hyperparameter settings for the DyGIE++ experiments in Table~\ref{tab:hyper-dygie}.
Experiments for DyGIE++ were run on a NVIDIA GeForce RTX 2080 Ti machine with support for 4 GPUs.

\begin{table}[t]
    \centering
    \begin{tabular}{lr}
        \toprule
        PLM & BERT-Large \\
        Training Batch Size & $6$ \\
        Eval Batch Size & $12$ \\
        Learning Rate & $2 \times 10^{-5}$ \\
        \# Training Epochs & $200$ \\
        Evaluation per \# Epoch & $1$ \\
        Max Sequence Length & $175$ \\
        \# Beams & 8 \\
        \bottomrule
    \end{tabular}
    \caption{Hyperparameter details for DyGIE++ model.}
    \label{tab:hyper-dygie}
\end{table}

\subsection{OneIE}

We report the hyperparameter settings for the OneIE experiments in Table~\ref{tab:hyper-oneie}.
Experiments for OneIE were run on a NVIDIA GeForce RTX 2080 Ti machine with support for 4 GPUs.

\begin{table}[t]
    \centering
    \begin{tabular}{lr}
        \toprule
        PLM & BERT-Large \\
        Training Batch Size & $6$ \\
        Eval Batch Size & $12$ \\
        Learning Rate & $1 \times 10^{-5}$ \\
        \# Training Epochs & $150$ \\
        Evaluation per \# Epoch & $1$ \\
        Max Sequence Length & $175$ \\
        \# Beams & 8 \\
        \bottomrule
    \end{tabular}
    \caption{Hyperparameter details for OneIE model.}
    \label{tab:hyper-oneie}
\end{table}

\subsection{Query\&Extract}

We report the hyperparameter settings for the Query\&Extract experiments in Table~\ref{tab:hyper-qe}.
Experiments for OneIE were run on an NVIDIA GeForce RTX 2080 Ti machine with support for 4 GPUs.

\begin{table}[t]
    \centering
    \begin{tabular}{lr}
        \toprule
        PLM & BERT-Large \\
        Training Batch Size & $16$ \\
        Eval Batch Size & $16$ \\
        Learning Rate & $5 \times 10^{-5}$ \\
        Weight Decay & $0.001$ \\
        \# Training Epochs & $5$ \\
        Evaluation per \# Epoch & $10$ \\
        Entity Embedding Size & $100$ \\
        \bottomrule
    \end{tabular}
    \caption{Hyperparameter details for Query\&Extract model.}
    \label{tab:hyper-qe}
\end{table}

\subsection{TE}

We use the original SRL engine and model provided in the repo for running the TE model. Since there was no training, we do not change any hyperparameters.



\begin{figure*}[t]
    \centering
    \includegraphics[width=\textwidth,height=\textwidth]{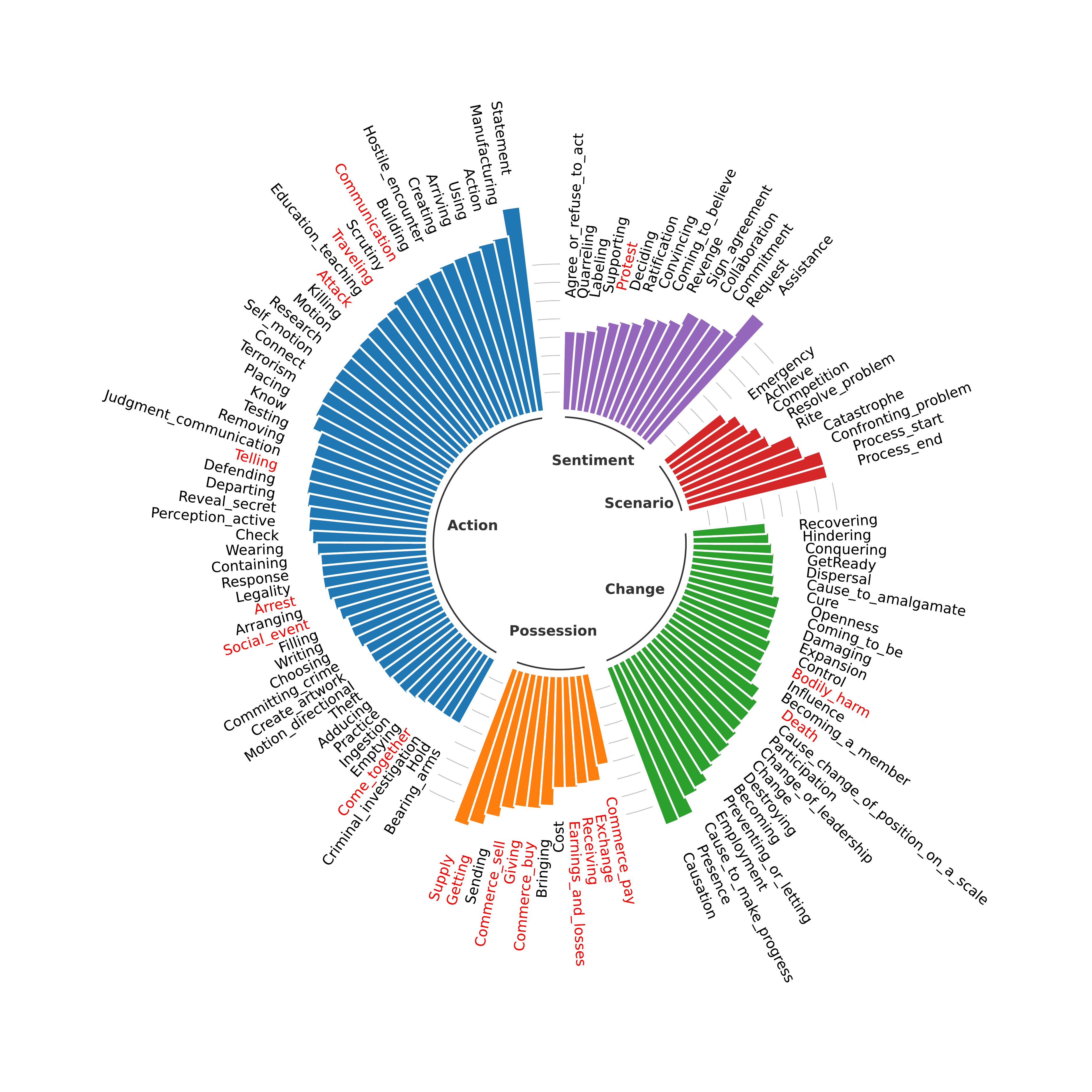}
    \caption{Circular bar plot for the various event types present in the \dataName dataset organized into abstract event types. The height of each bar is proportional to the number of event mentions for that event (height is in log-scale). Bar labels colored in \textcolor{red}{red} are the set of overlapping event types mapped from the ACE dataset.
    }
    \label{fig:event-organization}
\end{figure*}

\section{Complete Results}
\label{sec:all-results}

\begin{figure*}
    \centering
    \includegraphics[width=0.98\textwidth]{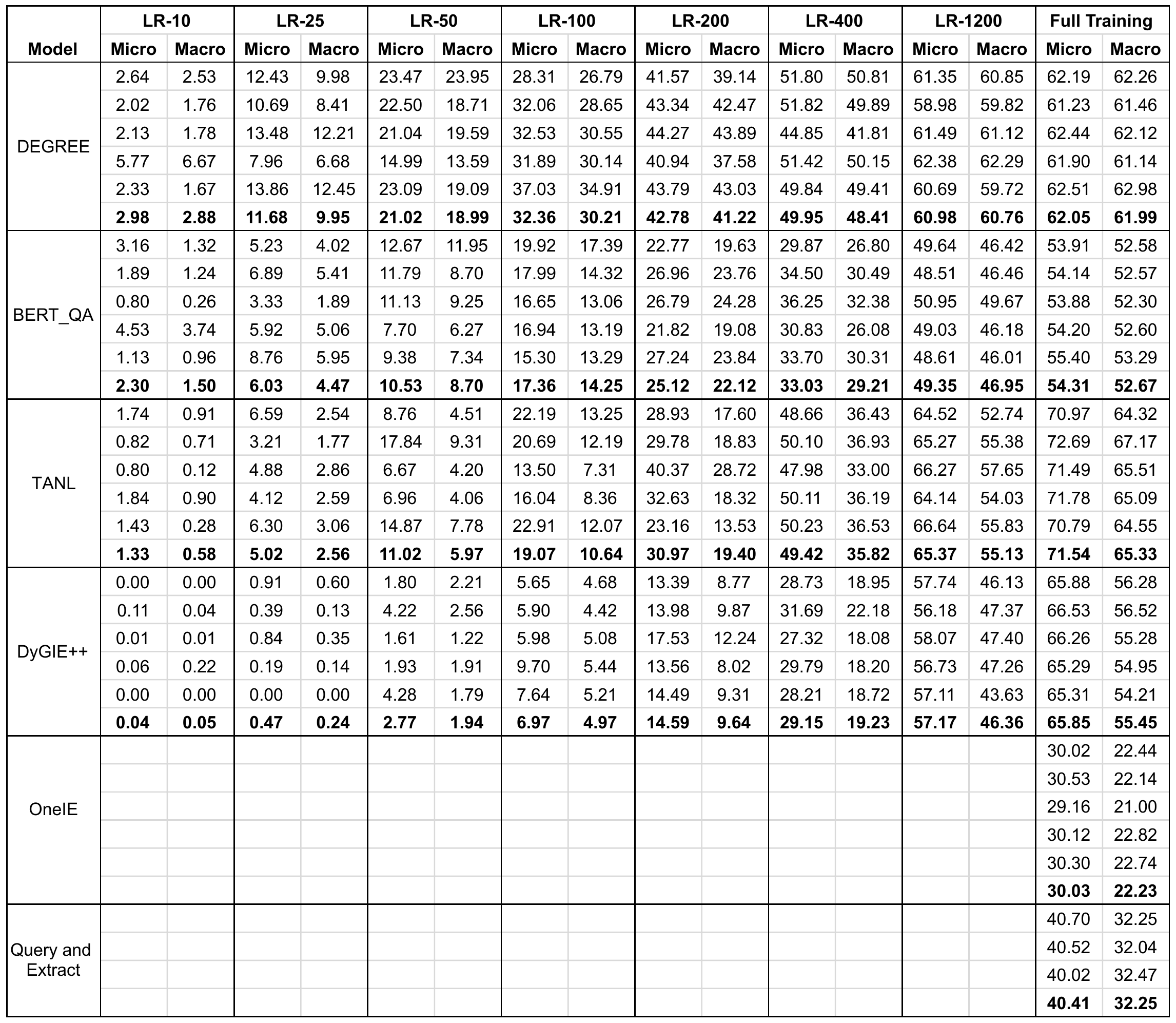}
    \caption{Complete set of results of the 5 different runs for all models for the low resource test suite. Here Micro is the micro F1 score and Macro is the macro F1 score. LR-XX denotes low resource with XX training mentions.}
    \label{fig:lr-complete}
\end{figure*}

\begin{figure*}
    \centering
    \includegraphics[width=0.98\textwidth]{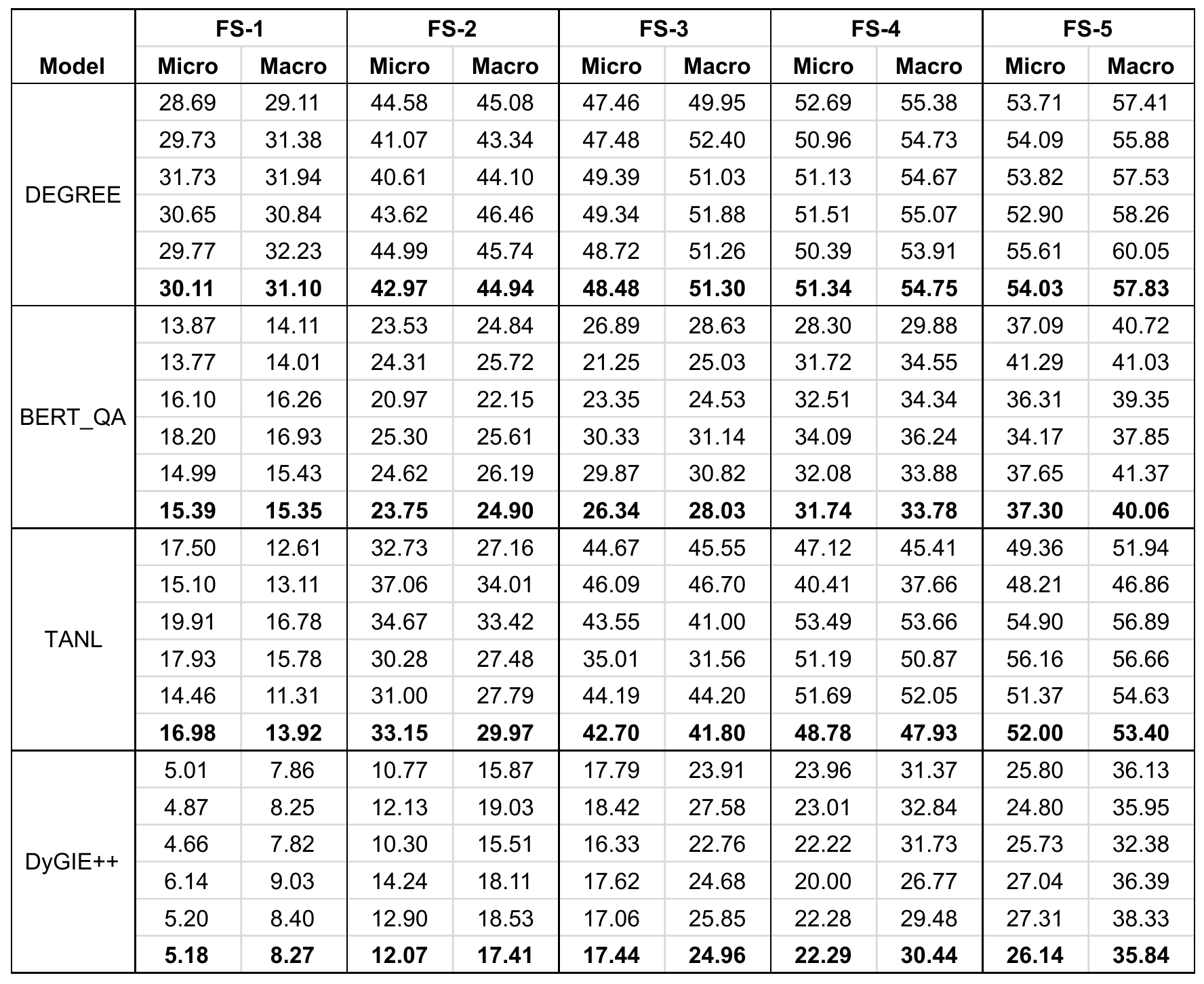}
    \caption{Complete set of results of the 5 different runs for all models for the few-shot test suite. Here Micro is the micro F1 score and Macro is the macro F1 score. FS-X denotes few-shot with X training mentions per event.}
    \label{fig:fs-complete}
\end{figure*}

\begin{figure*}
    \centering
    \includegraphics[width=0.98\textwidth]{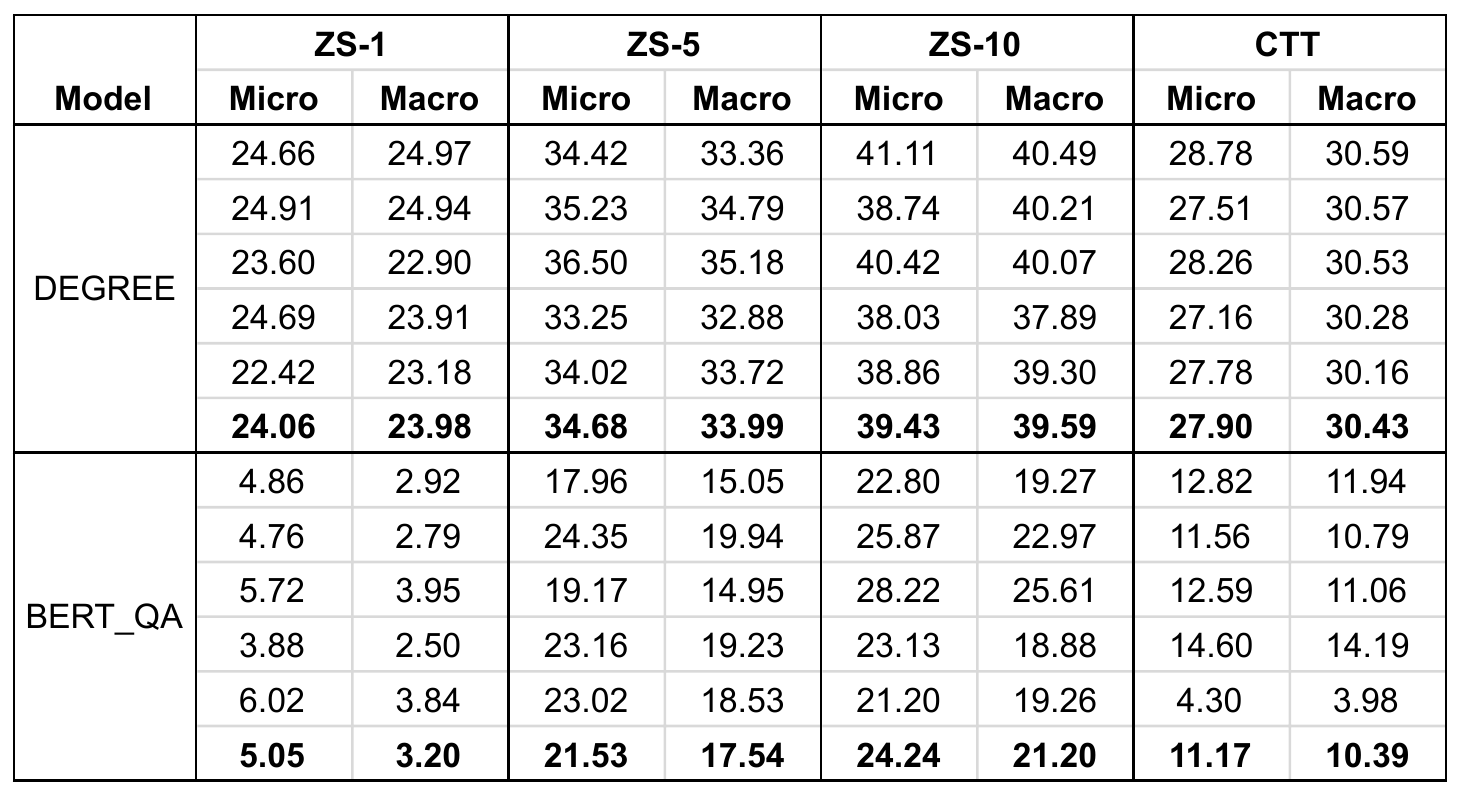}
    \caption{Complete set of results of the 5 different runs for all models for the zero-shot (ZS) and cross-type transfer (CTT) test suite. Here Micro is the micro F1 score and Macro is the macro F1 score. ZS-X denotes zero-shot with X training events.}
    \label{fig:unseen-complete}
\end{figure*}

In this section, we present the exhaustive set of results for each of the runs for the different benchmarking suites.
We show the results for the low resource and few-shot setting are shown in Figures~\ref{fig:lr-complete} and ~\ref{fig:fs-complete} respectively.
Figure~\ref{fig:unseen-complete} displays the results for the zero-shot and cross-type transfer settings.

\end{document}